\documentclass{article}

\PassOptionsToPackage{numbers, compress}{natbib}
\usepackage[final]{neurips_2020}

\usepackage[utf8]{inputenc} % allow utf-8 input
\usepackage[T1]{fontenc}    % use 8-bit T1 fonts
\usepackage{hyperref}       % hyperlinks
\usepackage{url}            % simple URL typesetting
\usepackage{booktabs}       % professional-quality tables
\usepackage{amsfonts}       % blackboard math symbols
\usepackage{nicefrac}       % compact symbols for 1/2, etc.
\usepackage{microtype}      % microtypography
\usepackage{graphicx}
\usepackage{array}
\usepackage{multirow}
\usepackage{amsmath}
\usepackage{amssymb}
\usepackage{float} 
\usepackage{subfigure}
\usepackage{wrapfig}
\usepackage[ruled]{algorithm2e}

\makeatletter
\newcommand{\rmnum}[1]{\romannumeral #1}
\newcommand{\Rmnum}[1]{\expandafter\@slowromancap\romannumeral #1@}
\makeatother

\title{Learning Multi-Agent Coordination for Enhancing Target Coverage in Directional Sensor Networks}

\author{%
  Jing Xu\textsuperscript{* \rm 1, 6}, Fangwei Zhong\textsuperscript{* \rm 2, 3, 5}, Yizhou Wang\textsuperscript{\rm 2, 4} \\
  \textsuperscript{*} indicates equal contribution\\
  \textsuperscript{1} Center for Data Science, Peking University 
\\
\textsuperscript{2} Dept. of  Computer Science, Peking University\\
\textsuperscript{3} Adv. Inst. of Info. Tech, Peking University\\ 
\textsuperscript{4} Center on Frontiers of Computing Studies, Peking University\\
\textsuperscript{5} Advanced Innovation Center For Future Visual Entertainment,  Beijing Film Academy\\
\textsuperscript{6} Deepwise AI Lab\\
  \texttt{jing.xu@pku.edu.cn, zfw@pku.edu.cn, yizhou.wang@pku.edu.cn} \\
}

\begin{document}
\maketitle
\begin{abstract}
%Human usually achieve high level coordination by dynamically choosing some specific targets and executing primitive actions to perform these skills, such as Predator-Prey and tracking problem. [https://arxiv.org/pdf/1912.03558.pdf]
Maximum target coverage by adjusting the orientation of distributed sensors is an important problem in directional sensor networks (DSNs). This problem is challenging as the targets usually move randomly but the coverage range of sensors is limited in angle and distance. Thus, it is required to coordinate sensors to get ideal target coverage with low power consumption, e.g. no missing targets or reducing redundant coverage.
To realize this, we propose a Hierarchical Target-oriented Multi-Agent Coordination (HiT-MAC), which decomposes the target coverage problem into two-level tasks: targets assignment by a \emph{coordinator} and tracking assigned targets by \emph{executors}.
Specifically, the coordinator periodically monitors the environment globally and allocates targets to each executor. In turn, the executor only needs to track its assigned targets.
To effectively learn the HiT-MAC by reinforcement learning, we further introduce a bunch of practical methods, including a self-attention module, marginal contribution approximation for the coordinator, goal-conditional observation filter for the executor, etc.
Empirical results demonstrate the advantage of HiT-MAC in coverage rate, learning efficiency, and scalability, comparing to baselines.
We also conduct an ablative analysis on the effectiveness of the introduced components in the framework.

\end{abstract}

\section{Introduction}
% 问题背景描述,场景规则
We study the target coverage problem in Directional Sensor Networks (DSNs). In DSNs, every node is equipped with a "directional" sensor, which perceives a physical phenomenon in a specific orientation. 
Cameras, radars, and infrared sensors are typical examples of directional sensors. In some real-world applications, the sensors in DSNs are required to dynamically adjust their own orientation to track mobile targets, such as automatically capturing sports game videos\footnote{https://playsight.com/automatic-production/}, actively tracking interesting objects~\cite{li2020pose}.
To realize these applications, the target coverage acts as a crucial point, which puts emphasis on \textbf{how to cover the maximum number of targets with the finite number of directional sensors}.
It is challenging as the targets usually move randomly but the locations of sensors are fixed. Meanwhile, the coverage range for sensors is limited in angle and distance.
To do this, it is required to collaboratively adjust the orientation of each sensor in DSNs by a multi-agent system to cover targets.
% To do this, the system is required to localize the targets, predict the future trajectories, and control the sensors. 
In practice, the multi-agent system for DSNs should : 1) accomplish the global task via multi-agent collaboration/coordination 2) be of good generalization to different environments 3) be low-cost in communication and power consumption.

In this paper, we are interested in building such a multi-agent collaborative system via multi-agent reinforcement learning (MARL), where the agents are learned by trial and error.
The simplest way is to build a centralized controller to globally observe and control the DSNs simultaneously.
And we can formulate it as a single-agent RL problem and directly optimize the controller by the off-the-shelf algorithms ~\cite{mnih2016asynchronous, schulman2017proximal}.
% However, the joint action space will dramatically increase with the number of sensors, leading to inefficient exploration. 
However, it is usually infeasible in real-world scenarios.
It is because that the system highly relies on real-time communication between the controller and sensors.
Moreover, it is hard to further extend the system for large scale networks, as the computational cost in the server will be dramatically expanded with the increasing agent numbers.
Recently, the RL community has taken great efforts on learning a fully decentralized multi-agent collaboration~\cite{lowe2017multi,rashid2018qmix,foerster2018counterfactual} for various applications, e.g. playing real-time strategy games~\cite{vinyals2019alphastar}, controlling traffic light ~\cite{chu2019multi}, self-organizing swarm system\cite{hung2016q}.
In a decentralized system, each agent runs individually, which observe the environment by themselves and exchange their information by peer-to-peer communication.
Such a decentralized system could run on a large scale multi-agent system and be low-cost on communication (even without communication).
But in most cases, the distributed policy is unstable and difficult to learn, as they usually affect others leading to a non-stationary environment.
Even though this issue has been mitigated by the recent centralized training and decentralized execution methods~\cite{lowe2017multi,rashid2018qmix,foerster2018counterfactual}, a remaining open challenge is how to effectively train a centralized critic to decompose the global reward to each agent for learning the optimal distributed policy, i.e. multi-agent credit assignment problem~\cite{nguyen2018credit,2019arXiv190705707W}.
% determine the contribution of each agent to the teamwork while training the centralized critic with a common team reward, i.e. multi-agent credit assignment problem~\cite{nguyen2018credit,2019arXiv190705707W}.
% It is important for the multi-agent learning, especially when agents share a common team reward, as 
% It is necessary to effectively find an optimal policy for each when they only share a team reward.
To this end, we are motivated to explore a feasible solution to combine the advantages of above methods to learn a multi-agent system for the target coverage problem effectively.

\begin{figure*}[b]
\vspace{-0.5cm}
  \centering
    \subfigure[]{
    \label{hierarchy}
    \vspace{-0.6cm}
    \includegraphics[width=0.45\linewidth]{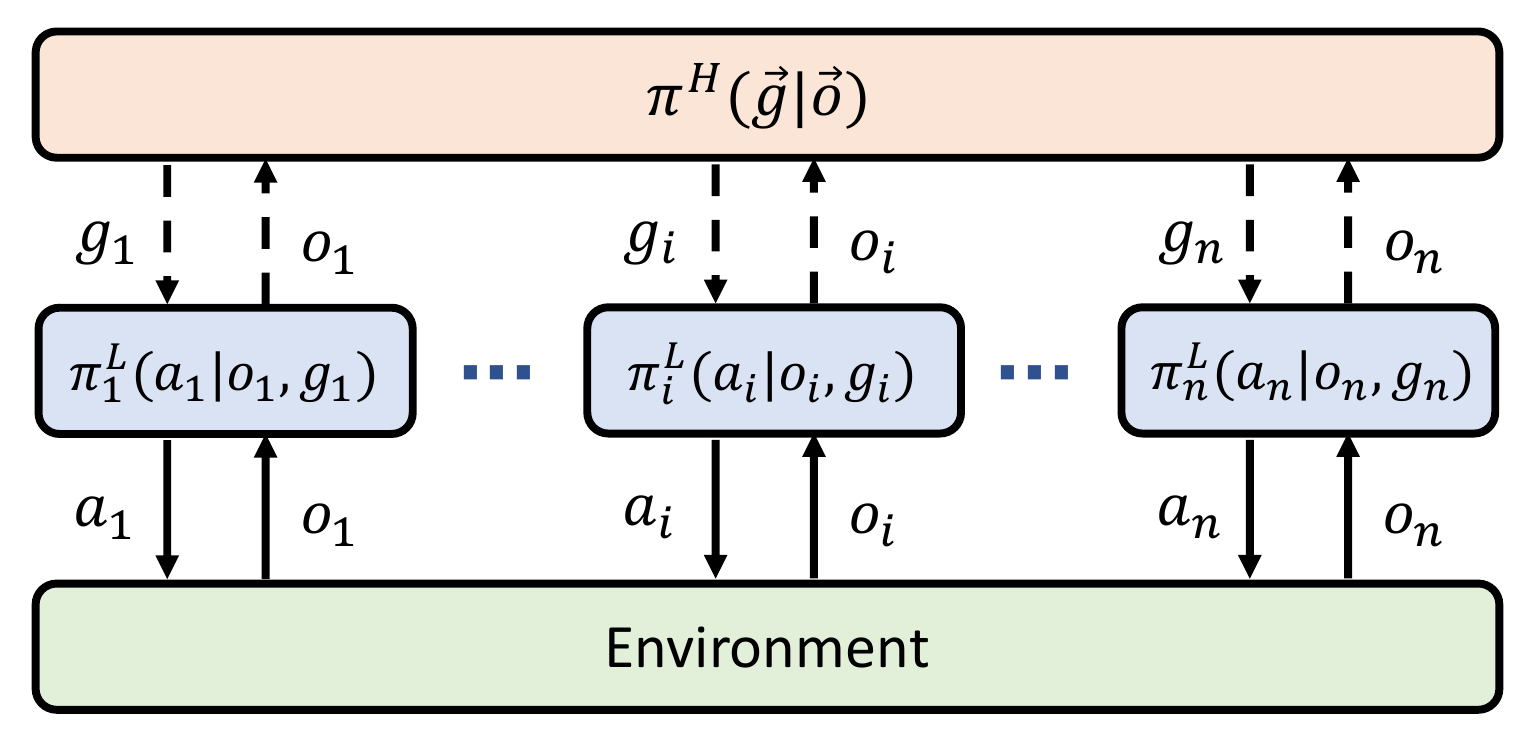}
    }
    \subfigure[]{
    \label{network}
    \includegraphics[width=0.45\linewidth]{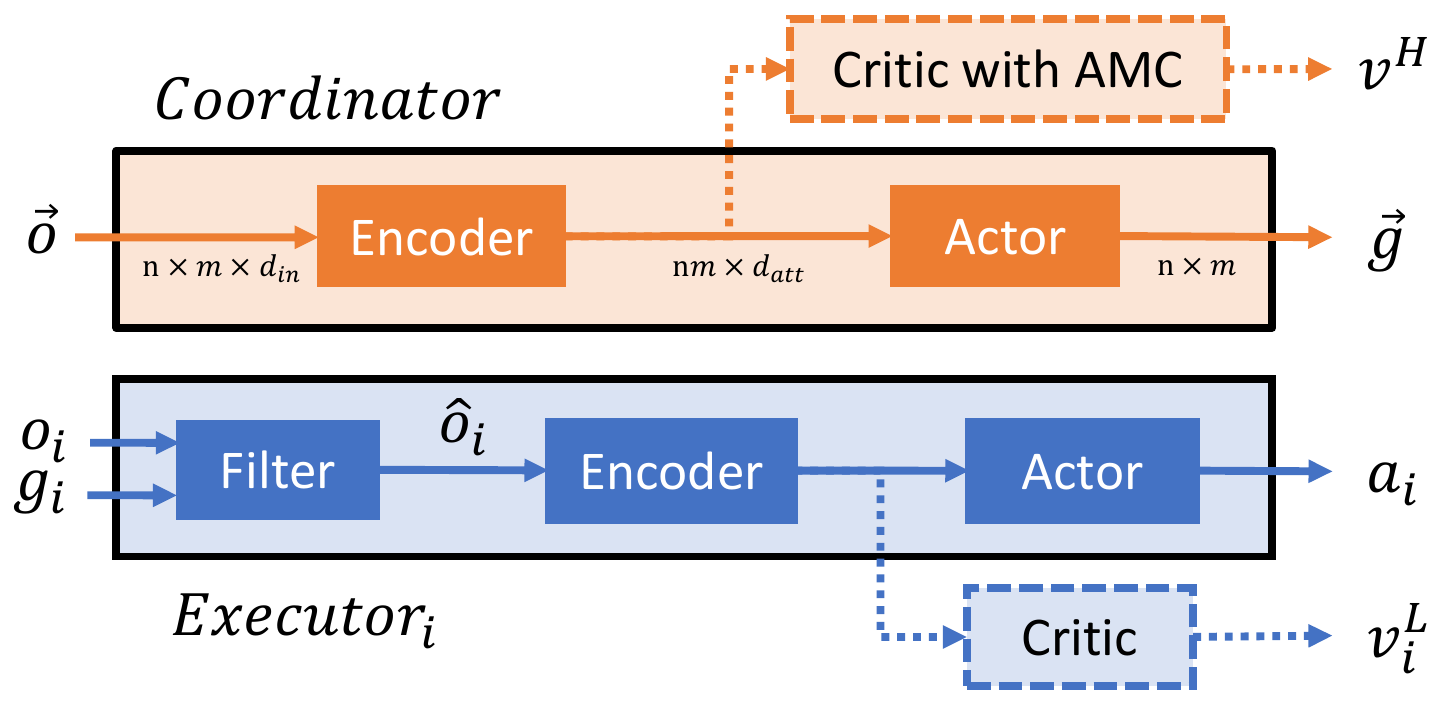}
    }
\caption{An overview of the HiT-MAC framework. Fig.~\ref{hierarchy} is the two-level hierarchy of HiT-MAC. 
  Periodically (every $k$ steps), the high-level policy (\emph{coordinator}) $\pi^H(\vec{g}|\vec{o})$ collects joint observation $\vec{o}=(o_1,\dots,o_n)$ from sensors and distributes target-oriented goal $g_i$ to each low-level policy (\emph{executor}).
  In turn, the executor $\pi_i^L(a_i|o_i, g_i)$ directly interacts with the environment to track its own targets.
  The observation $o_i$ describes the spatial relation between sensor $i$ and targets.
  The goal $g_i$ allocates the targets to be followed by the executor $i$.
  Note that the solid line and the dashed line are executed at every step and every $k$ steps respectively.
  Fig.~\ref{network} is the details of the coordinator and executor. The critics for them are only used while training the networks. Please refer to Sec.~\ref{HiT-MAC} for more details.
  }
  \label{framework}
\vspace{-0.5cm}
\end{figure*}

We propose a Hierarchical Target-oriented Multi-agent Coordination framework (HiT-MAC) for the target coverage problem, inspired by the recent success in Hierarchical Reinforcement Learning (HRL)~\cite{vezhnevets2017feudal, kong2017revisiting, li2019hierarchical}. This framework is a two-level hierarchy, composed of a centralized \emph{coordinator} (high-level policy) and a number of distributed \emph{executors} (low-level policy), shown as Fig.~\ref{framework}.
While running, (a) the \emph{coordinator} collects the observations from executors and allocates goals (a set of targets to track) for each executor, and (b) each \emph{executor} individually takes primitive actions to complete the given goal for $k$ time steps, i.e. tracking the assigned targets. After the $k$ steps execution, the coordinator is activated again. Then, steps a and b iterate.
% Periodically, the coordinator is activated to collect the joint observations and schedule the sub-task for each executor. 
% Concretely, the coordinator specifies the targets for executors to track, respectively.
% Continually, each sensor is rotated by an executor individually to cover the assigned targets.
% Continually, each sensor is rotated by an executor individually, which focuses on tracking a specific set of targets by optimizing a goal-conditioned reward.
In this way, the target coverage problem in DSNs is decomposed into two sub-tasks at different temporal scales. Both coordinator and executors can be trained by the modern single-agent reinforcement learning method (e.g. A3C~\cite{mnih2016asynchronous}) to maximize expected future team reward (coordinator) and goal-conditioned rewards (executors), respectively.
Specifically, the team reward is factored by the coverage rate; the goal-conditioned reward is about the performance of a sensor to track the selected targets, measured by the relative angle among sensor and target. 
% targets assignment and tracking targets by rotating sensors
% Hence, the multi-agent credit assignment problem is also mitigated by converting to the target assignment problem in the coordinator.
So, it can also be considered as the cooperation between the coordinator and executors.

To implement a scalable HiT-MAC, there are two challenges to overcome: (1) For the coordinator, how to learn a policy to handle the assignment among variable numbers of sensors and targets? (2) For the executor, how to train a robust policy that could perform well in any possible cases, e.g. given different target combinations?
Hence, we employ a battery of practical methods to address these challenges. Specifically, we adopt the self-attention module to handle variable input size and generate a order-invariant representation. 
We estimate values by approximating the marginal contribution (AMC) of each pair of the sensor-target assignments.
With this, the critic could better estimate and decompose the team value in a more accurate way, which guides to a more effective coordination policy.
For the executor, we further introduce a goal-conditioned filter to remove the observation of the irrelative targets and a goal generation strategy for training.

We demonstrated the effectiveness of our approach in a simulator, comparing with the state-of-the-art MARL methods, and a heuristic method.
% Empirical results in a numerical simulator demonstrate the effectiveness of the proposed framework, while it shows that HMCM outperforms the state-of-the-art MARL methods, and conventional optimization-based method. 
To be specific, our method achieves the highest coverage ratio and fastest convergence in the case of $4$ sensors and $5$ targets. We also validate the good transferability of HiT-MAC in environments with different numbers of sensors ($2\sim 6$) and targets($3\sim 7$).
Besides, we also conduct an ablation study to analyze the contribution of each key component.

Our contributions can be summarized in three-folds:
\begin{itemize}
    \item We study the target coverage problem in DSNs and propose a Hierarchical Target-oriented Multi-agent Coordination framework  (HiT-MAC) for it. To the best of our knowledge, it is the first hierarchical reinforcement learning method for this problem.
    \item A bunch of practical methods is introduced to effectively learn a generalizable HiT-MAC, including a self-attention module, marginal contribution approximation, goal-conditioned filter, and so on.
    \item We release a numerical simulator to mimic the real scenario and conduct experiments in the environments to illustrate the effectiveness of the introduced framework.
\end{itemize}

\section{Preliminary} % Background and 
\textbf{Problem Definition.}
% power cost
The target coverage problem considers how to use a number of active sensors to continuously cover maximum number of targets.
In this case, there are $n$ sensors and $m$ mobile targets in the environment. 
Sensors are randomly placed in the environment with limited coverage range.
The targets randomly walk around the environment.
A target is covered by the sensor network, once it is monitored by at least one sensor.
The orientation of the sensor is adjustable, but the changing angle at each step is restricted considering the physical constraints.
Besides, considering the efficiency problem, every movement will take additional cost in power.

\textbf{Dec-POMDPs.}
It is natural to formulate the target coverage problem in $n$ sensors networks as a Dec-POMDPs~\cite{bernstein2002complexity}). It is governed by the tuple $\langle N, S, \{A_i\}_{i \in N}, \{O_i\}_{i \in N}, R, Pr, Z \rangle$ where: $N$ is a set of $n$ agents, indexed by $\{1, 2,..., n\}$; $S$ is a set of world states; $A_i$ is a set of primitive actions available for agent $i$, and forming joint actions $\vec{a_t} = (a_{1,t},...,a_{n,t})$ with others; $O_i$ is the observation space for agent $i$, and its local observation $o_{i,t} \in O_i$ is drawn from observation function $Z(o_{i,t}| s_{t}, \vec{a}_{t})$; $R: S \rightarrow R$ is the team reward function, shared among agents; $Pr: S \times A_1 \times ... A_n \times S \rightarrow [0, 1]$ defines the transition probabilities between states over joint actions.
Notably, the subscript $t \in \{1, 2,...\}$ denotes the time step.
% $Z$ defines observation probabilities over joint actions and joint observations.
% The subscript $i \in \{1, 2,..., n\}$ denotes the index of each agent.
At each step, each agent acquires observation $o_{i,t}$ and takes an action $a_{i,t}$ based on its policy $\pi_{i}(a_{i,t}|o_{i,t})$.
Influenced by the joint action $\vec{a}_t$, the state $s_t$ is updated to a new state $s_{t+1}$ according to $Pr(s_{t+1}|s_t, \vec{a}_t)$.
Meanwhile, the agent $i$ receives the next observation $o_{i, t+1}$ and the team reward $r_{t+1}=R(s_{t+1})$.
% and joint observation $\vec{o}_{t+1}$ are drawn from the environment transition function $P(s_{t+1} |s_t, \vec{a}_t)$ and the observation function $Z(\vec{o}_{t+1}| s_{t+1}, \vec{a}_{t})$.
% Meanwhile, the agents share a global reward $R_{g, t} = R(s_t, \vec{a_t})$ to incentive a better strategy. The global reward is based on the coverage rate of the $M$ targets $\sum_{j=1}^{m} D_{j,t}$. $D_{j,t}$ indicates whether the target $j$ is covered by the sensor networks.
% The policy of each agent, $\pi_{i}(a_{i,t}|o_{i,t})$, is a distribution over action $a_{i,t}$ conditioned on its observation $o_{i,t}$.
For the cooperative multi-agent task, the ultimate goal is to optimize the joint policy $<\pi_{1},...,\pi_{n}>$ to maximize the $\gamma$ discounted accumulated reward with time horizon $T$: 
$\mathbb{E}_{\vec{a}_t\sim<\pi_{1},...,\pi_{n}>} \left[ \sum_{t=1}^{T} \gamma^t r_{t} \right]$.
% $\mathbb{E}_{\vec{a_t}\sim<\pi_{1},...,\pi_{n}>} \left[ \sum_{t=1}^{T} \lambda r_{t} \right]$
% where $N, S, O, A, R, Pr, Z$ denote a set of $n$ agents, state space, observation space, action space, reward function, transition function, and observation function respectively.
% Note that a group of $m$ targets will be specifically considered, as it is the key element for the observation and the reward function in the coverage problem.
% In the case of partial observation, we have the observation $o_{i,t} = o_{i,t}(s_{t})$, where $o_{i,t} \in O_{i}$, $s_t \in S$.
% Focusing the sensor-target relation, the observation can be written as: $o_{i}=<d_{i,1}, d_{i,2}, ..., d_{i,m}>$.
% Note that $d_{i,j}$ is the relation information between sensor $i$ and target $j$, e.g. distance, angle, etc.
% There are the joint observation $\vec{o}_t = <o_{1,t},...,o_{n,t}>$  and output the joint action $\vec{a_t} = <a_{1,t},...,a_{n,t}>$ in the multi-agent system.

% Generally, one potential solution is directly employing the off-the-shelf multi-agent reinforcement learning method.
% However, we find that they are unstable while learning and cannot perform well.

\textbf{Hierarchical MMDPs.}
Considering the hierarchical structure of the coverage problem, we decompose it into two tasks: high-level coordination and low-level execution. 
The high-level agent (\emph{coordinator}) focuses on coordinating $n$ low-level agents (\emph{executors}) in the long-term to maximize the accumulated team reward $\sum_{t=1}^{T} \gamma^t r_{t}$.
To do it, the \emph{coordinator} $\pi^H(\vec{g}_t|\vec{o}_t)$ distributes goals $\vec{g}_t=(g_{0,t}, g_{1,t},\dots, g_{n,t})$ to the \emph{executors}, based on the joint observation $\vec{o}_t$ ( collected from executors).
After receiving the goal $g_{i,t}$ at time step $t$, the executor $i$ locally accomplishes the goal for $k$ steps, i.e., maximizing the cumulative goal-conditioned reward $r^L_{i,t} = R^L(s_t, g_{i,t})$, by continuously taking primitive actions $a_i$ based on the policy $\pi^L_i(a_{i,t}|o_{i,t})$.
% Once the \emph{executor} $i$ receives the goal $g_{i,t}$ at time step $t$, the low-level MDPs update $k$ steps continuously by executing primitive actions $a_i$.
% The low-level policy $\pi^L_i(a_{i,t}|o_{i,t})$ aims at locally accomplish the goal $g_{i,t}$ by maximizing the cumulative goal-conditioned reward $r^L_{i,t} = R^L(s_t, g_{i,t})$, where $R^L$ is an local reward function to evaluates the achievement of the goal.
%Since the goal $\vec{g}_t$ for low-level transition would last for $k>1$ steps until the next update,
Since the coordinator interacts with executors every $k>1$ steps, 
the high-level transition could be regarded as a semi-MDP~\cite{sutton1999between}.
And the executors still run on a decentralized style as the Dec-POMDP. Differently, the reward function and policy of each are directed by its goal $g_{i, t}$ introduced in the hierarchy.
Thus, the semi-MDP and Dec-POMDPs form a two-level hierarchy for multi-agent decision making, referring as a hierarchical Multi-agent MDPs(HMMDPs).

\textbf{Attention Modules.}
Attention modules~\cite{vaswani2017attention,cheng2016long} have attracted intense interest due to the great capability in a lot of different tasks~\cite{xu2015show,anderson2018bottom,vinyals2015pointer}.
Furtherly, the self-attention module can handle variably-sized inputs in an order-invariant way. 
In the paper, we adopt the scaled dot-product attention~\cite{vaswani2017attention}. 
Specifically, the matrix of output $\mathbf{H}$ is a weighted sum of the values, which is computed as:
\begin{equation}
\label{attention}
    \mathbf{H} = Att(\mathbf{Q},\mathbf{K},\mathbf{V}) = softmax(\frac{\mathbf{Q}\mathbf{K}^T}{\sqrt{d_k}})\odot \mathbf{V}
\end{equation}
where $d_k$ is the dimension of a key; the matrix $\mathbf{K}$, $\mathbf{Q}$, $\mathbf{V}$ are the keys, queries, and values, transformed from input matrix $\mathbf{X}$ by parameter matrices $\mathbf{W}_q$, $\mathbf{W}_k$, $\mathbf{W}_v$. They are computed as:
\begin{equation}
    \mathbf{Q}=tanh(\mathbf{W}_q \mathbf{X}), K=tanh(\mathbf{W}_k \mathbf{X}), \mathbf{V}=tanh(\mathbf{W}_v \mathbf{X})
    \label{att-2}
\end{equation}
% where $W_q$, $W_k$, $W_v$ are parameter matrices transforming $X$ into queries, keys, and values separately.
The context feature $C=\sum_{i=1}^N h_i$ summarizes elements in \textbf{H} in an additive way, where $h_i$ and $N$ are the element and the total number of elements in \textbf{H}.

\textbf{Approximate Marginal Contribution.} 
In the cooperative game, the marginal contribution $\varphi_{C,i}(s)$ of the member $i$ in a coalition $C$ is the incremental value brought by the joining of member $i$.
% amount by which the overall value created would shrink if the member in question were to leave the cooperative game.
%that is from the game theory \cite{Osborne1994A}, 
Formally, it is $\varphi_{C,i}(v)=v(C\cup\{i\})-v(C)$, where $v(\cdot)$ represents the value of a coalition.
In a $N$ player setting, 
Shapley value\cite{Shapley1953A} measures the average of marginal contributions of member $i$ in all possible coalitions, written as $\sum_{C\in N\backslash i} \frac{|C| !(N-|C|-1) !}{N !} \varphi_{C,i}(v)$.
$N\backslash i$ denotes the subset of $N$ consisting of all the players except member $i$.
% In a $N$ player setting, to better estimate The contribution of the $i$-th member $\varphi_{i}(s)$, Shapley value\cite{Shapley1953A} takes the average of marginal contributions of all possible coalitions , written as $\sum_{C\in N\backslash i} \frac{|C| !(N-|C|-1) !}{N !} \varphi_{C,i}(v)$, where $N\backslash i$ denote the subset of $N$ consisting of all the players except member $i$.
Thus, the contributions made by every member can be calculated, once all the sub-coalition contributions $v(C)$ are given.
However, it is infeasible to calculate it in practice, as the number of all possible coalitions will be expanded with increasing members $N$, which causes the computational catastrophe.
% However, the sub-coalition contribution is hard to calculate in practice, while the grand coalition contribution can be measured by the global reward. 
Hence, ~\cite{2019arXiv190705707W} introduced a method to approximate the marginal contribution by deep neural networks.
In this paper, we approximate the marginal contribution of each pair of sensor-target assignments by neural network for learning a coordinator effectively, rather than estimate the marginal contribution of each player.

%the marginal contribution of a particular player is the amount by which the overall value created would shrink if the player in question were to leave the game.

\section{Hierarchical Target-oriented Multi-Agent Coordination}
\label{HiT-MAC}

% \begin{figure}
%   \centering
%   \includegraphics[width=\linewidth]{figures/framework1.pdf}
%   \vspace{-0.4cm}
%   \caption{An overview of HiT-MAC. \textbf{Left}: the two-level hierarchy of HiT-MAC. 
%   The high-level policy (\emph{coordinator}) $\pi^H(\vec{g}|\vec{o})$ monitors the environment by joint observation $\vec{o}=(o_1,\dots,o_n)$ and distributes target-oriented goals $\vec{g}=(g_1,\dots,g_n)$ to each low-level policy (\emph{executor}). 
%   In turn, the executor $\pi_i^L(a_i|o_i, g_i)$ directly interact with the environment to track its own targets.
%   The observation $o_i$ describes the relation between sensor $i$ and all targets.
%   The $g_i$ points out which targets should the executor $i$ focus on.
%   Note that the solid line and the dashed line in the left indicate that it is executed every step and every $k$ steps respectively.
%   \textbf{Right}: the details of the coordinator and executor.
%   The critics are only used while training the networks.
%   Please refer to Sec.~\ref{HiT-MAC} for more details.
%   }
%   \label{framework}
%   \vspace{-0.3cm}
% \end{figure}

 Hierarchical Target-oriented Multi-Agent Coordination (HiT-MAC) is a two-level hierarchy, consisting of a coordinator (high-level policy) and $n$ executors (low-level policy), shown as Fig.~\ref{framework}. The coordinator and executors respectively follow the semi-MDP and goal-conditioned Dec-POMDPs in HMMDPs. Periodically, the coordinator aggregates the observations $\vec{o}=(o_1, o_2,\dots, o_n)$ from the executors and distributes a target-oriented goal $\vec{g}=(g_1, g_2,\dots, g_n)$ to them.
 After receiving $g_i$, the executor $i$ will minimize the average angle error to the assigned targets by rotation for $k$ steps based on its policy $\pi_i^L(a_i|o_i, g_i)$.
 The framework is target-oriented in three-folds: 
 1) the observation $o_i$ describes the spatial relations among sensor $i$ and all targets $M$ in the environment;
 2) the $g_i$ explicitly identifies a subset of targets $M_i \subseteq M$ for the executor $i$ to focus on.
 3) the rewards for both levels are highly dependent on the spatial relations among sensors and targets, i.e, the team reward is about the overall coverage rate of targets, the reward for the executor $i$ is about the average angle error between the executor $i$ and its assigned targets.
 
In the following, we will introduce the key ingredients for HiT-MAC in details. 
% For coordinator, we adopt an attention-based encoder and introduce a relation-oriented AMC for the critic. For every executor $i$, a goal-conditioned observation filter and a pretraining strategy are employed.

\subsection{Coordinator: Assigning Targets to Executors}
The coordinator seeks to learn an optimal policy $\pi^{H *}(\vec{g}|\vec{o})$ that can maximize the cumulative team reward
%$\sum_{t=1}^{T} \gamma r_{t}$ 
by assigning appropriate targets $\{M_i\}_{i \in N}$ for each executor $i \in N$ to track.
% assigning a reasonable target set $\{M_i\}_{i \in N}$ to each executor respectively.
Note that the coordinator only runs periodically (every $k$ steps) to wait for the low-level execution and save the cost in communication and computation.
% Such coordination policy would lead to a more explicit cooperation strategy.
% we seek to learn representations which could capture the variably-sized and order-invariant set of targets in the environment. Meanwhile, the policy learning should be effective and generalizable to different scales.
% relation： 学更明确的合作关系
% \textbf{The overall network for the coordinator}

\textbf{Team reward function $r^H_t$} for the coordinator is equal to the target coverage rate $\frac{1}{m}\sum_{j=1}^{m} I_{j, t}$ if any target covered (Condition a). $I_{j,t}$ represents the covering state of target $j$ at time step $t$, where $1$ is being covered and $0$ is not. Notably, if none of targets is covered (Condition b), we will give an additional penalty in the reward.
The overall team reward is shown as following:
\begin{equation}
\label{eq:global_reward}
r^H_t = R(s_t) = 
\begin{cases}
\frac{1}{m}\sum_{j=1}^{m} I_{j, t} & (a) \\
-0.1 & (b)
\end{cases}
\end{equation}

The coordinator is implemented by building a deep neural network, which is composed of three parts: state encoder, actor, critic.
There are mainly two challenges to build the coordinator.
First, the shapes of the joint observation $\vec{o}$ and goal $\vec{g}$ depend on the number of sensors and targets in the environment. 
Second, it is inefficient to explore target-assignments only with a team reward directly, especially when the goal space is expanded with the increasing number of sensors and targets. 
Thus, the network should be capable of 1) handling the variably-sized input and output; 2) finding an effective approach for the critic to estimate values.

\textbf{State encoder} adopts the self-attention module to encode the joint observation $\vec{o} \in [o_{i,j}]_{n\times m}$ to an order-invariant representation $\textbf{H} \in \mathbb{R}_{n\times m \times d_{att}}$. 
% In the setting of DSNs, we focus on the sensor-target relation and the observation $o_{i}=<d_{i,1}, d_{i,2}, ..., d _{i,m}>$, given $m$ targets.
Note that $o_{i,j}$ is a $d_{in}$ dimensional vector, indicating the spatial relation between sensor $i$ and target $j$.
In our setting, $o_{i,j}=(i, j, \rho_{ij}, \alpha_{ij})$, where $\rho_{ij}$ and $\alpha_{ij}$ are the relative distance and angle respectively. Please refer to Sec.~\ref{env} for more details. 
% So, the joint observation $\vec{o} \in \mathbb{R}_{N\times M \times D_{in}}$.
To feed $\vec{o}$ into the attention module, we flatten it from $\mathbb{R}_{n\times m \times d_{in}}$ to $\mathbb{R}_{nm \times d_{in}}$, then encode it as $\mathbf{H}=Att(\mathbf{Q},\mathbf{K},\mathbf{V})$, where $\mathbf{Q}, \mathbf{K}, \mathbf{V}$ are derived from the flatten observation according to Eq.~\ref{att-2}.

\textbf{Actor} adaptively outputs the goal map $\vec{g} \in \mathbb{N}_{n\times m}$ according to $\mathbf{H} \in \mathbb{R}_{nm \times d_{att}}$. Firstly, we reshape \textbf{H} as $[n, m, d_{att}]$ again, and compute the probability $p_{ij}$ of each assignment by one fully connected layer, $p_{ij} = f_a(\mathbf{H}_{i,j})$.
Then, we sample the assignment $g_{i,j}$ by probability. $g_{i, j}$ is a binary value, indicating if let sensor $i$ to track target $j$.
At the end, the actor outputs the goal map $\vec{g}$ for executors, where $g_{i}=(g_{i,1},g_{i,2},...,g_{i,j})$ denotes the targets assignment for the sensor $i$.
% gather
 
\textbf{Critic} learns a value function, which is then used to update the actor's policy parameters in a direction of performance improvement.
Rather than directly estimating the global value by a neural network, we introduce an approximate marginal contribution (AMC) approach for learning the critic more efficiently.
% We regard the team value as the sum of the values of all potential sensor-target assignments, $v^H=\sum_{e=1}^{nm} \varphi_e$. 
Similar to most multi-agent cooperation problems, we deduce the individual contribution of each member to the team's success, referred as credit assignment.
% And then, the marginal contribution of each member can be approximated by the deep neural network.
% Usually, this is based on the assumption that the team reward is additively decomposable among members.
% However, we find that the team reward for the target coverage problem in is non-decomposable among agents.
Differently, we regard each sensor-target pair of the assignments, instead of the agent, as a member of the team.
It is because that the coordinator undertakes all the sensor-target assignments, which will directly affect global rewards (if the executors are perfect).
Identifying the contribution of each sensor-target assignment to the team reward will be beneficial for a reasonable and effective coordination policy, and such a policy leads to better cooperation among the executors.

% The critic is to comment the value of the current state, while we introduce the AMC module for the better credit assignment.
% Besides, the context vector $C_t=\sum_{i=1}^{N\times M}\psi^i_t$ can also be treated as the global feature for global reward estimation. But, the critic based on such global feature cannot work well which is shown in the ablation study. So, we introduce AMC to conduct a more effective credit assignment.

Inspired by~\cite{2019arXiv190705707W}, we approximate the marginal contribution of each assignment (assigning target $j$ to agent $i$) by neural network $\phi$. The input is $\textbf{H} \in \mathbb{R}_{nm \times d_{att}}$ from the state encoder. The length of $\textbf{H}$ is $l=nm$, then it can be regarded as a $l$-member cooperation.
So, the marginal contribution is approximated as $\varphi_{e} = \phi([\eta_{e},z_{e}])$. 
Here $[\cdot,\cdot]$ denotes concatenation, $\eta_{e}$ is the embedded feature of the sub-coalition $C_{e}=\{1, ...,e-1\}$ for the member $e$.
For example, if the grand coalition is $[z_1, z_2, z_3, z_4]$, then the $\eta_3$ is the context feature of $[z_1, z_2]$, which is used for computing the marginal contribution of the member $3$.
So, the credit assignment is conducted among all the pairwise sensor-target assignment in the coordinator as Alg.1.

\begin{algorithm}[H]
\label{AMC}
\caption{Estimate team value with AMC}
\LinesNumbered 
\KwIn{the state representation $\mathbf{H} \in \mathbb{R}_{nm \times d_{att}}$}
\KwOut{estimated global team value $v^H$}
Initialize the sub-coalition feature $\eta_1 = \mathbf{0}$\\
Given an attention module $Att'(\cdot)$ and a value network $\phi(\cdot)$\\
l = n*m \\
\For{e=1 to l}{
Compute the marginal contribution $\varphi_e = \phi([\eta_e, h_e])$, where $h_e$ is $e$-th element in $\mathbf{H}$\\
Compute element-wise features of the sub-coalition $\mathbf{H}' = Att'(\mathbf{H}[1:e]$)\\
Compute the embedded feature $\eta_{e+1} = \sum_{i=1}^e h'_i$, where $h'_i$ is the $i$-th element in $\mathbf{H}'$\\
}
The team value $v^H=\sum_{e=1}^{l} \varphi_e$\\

\end{algorithm}

Our AMC is conducted on value $v^H$, which is different from SQDDPG~\cite{2019arXiv190705707W}. It is because that AMC is conducted on value $Q$ in SQDDPG~\cite{2019arXiv190705707W}, which would introduce an extra assumption, i.e. the actions taken in $C$ should be the same as the ones in the coalition $C \cup \{i\}$, detailed in Appendix~\ref{app6.1}.
Our global value estimation is also different from the existing methods, like \cite{rashid2018qmix, sunehag2018value}, because ours refers the sub-coalition contribution to make a more confident estimation of the contribution from each member. Theoretically, the permutation of the coalition formation order should be sampled like the computation of Shapley Value\cite{Shapley1953A}. 
% However, we observe that AMC with permutation cannot outperform the one without permutation based on neural network. 
However, we observe that the permutation of hidden states is useless in our case. And the promotion caused by permutation is also not obvious enough shown in \cite{2019arXiv190705707W}.
So, we fix the order in the implementation, i.e. from $1$ to $l$.
% Then, the parameters of AMC will be updated by the gradients in the Actor-Critic framework:
% \begin{gather*}
%     d\theta_c\leftarrow d\theta_c + \tau \nabla_{\theta_c}(r+\gamma V(s';\theta_c)-V(s;\theta_c))^2\\
%     d\theta_a\leftarrow d\theta_a+ \tau (r+\gamma V(s';\theta_c)-V(s;\theta_c))\nabla_{\theta_a} log \pi(a; s,\theta_a);
% \end{gather*}

\subsection{Executor: Tracking Assigned Targets}
After receiving the goal $g_i$ from the coordinator, the executor $\pi^L_i(a_i|o_i, g_i)$ completes the goal-conditional task independently. In particular, the goal of executor $i$ is tracking a set of assigned targets $M_i$, i.e, minimize the average angle error to them.

For training, we further introduce a \textbf{goal-conditioned reward} $r^L_{i, t}(s_t, g_t)$ to evaluate the executor.
We score the tracking quality of the assigned targets based on the average relative angle, referring to Eq. \ref{eq:tracking_reward}.
We consider two conditions, which are (a) the target $j$ is in the coverage range of the sensor $i$, i.e. $ \rho_{ij, t} < \rho_{max} \& | \alpha_{ij, t} | < \alpha_{max} $; 
(b) target is outside of the range.
Here $\alpha_{max}$ is the maximum viewing angle of the sensor, $\alpha_{ij, t}$ is the relative angle from the front of the sensor to the target $j$.
% As for the target $j$, ideally, the sensor should orientate to it straightly. 
% We measure the power consumption $cost$,
% % If $a_{i,t}$ is $Stay$, $cost_{i,t}=0$. Otherwise, $cost_{i,t}=0.1$.
% according to the moved angle $|\delta_{i,t} - \delta_{i, t-1}|$ of the sensor's working direction $\delta_{i,t}$. 
\begin{equation}
\label{eq:tracking_reward}
r^L_{i, t} = \frac{1}{m_i}\sum_{j \in M_i} r_{i, j, t}-\beta cost_{i,t}
, 
r_{i, j, t} = 
\begin{cases}
1 - \frac{|\alpha_{ij, t}|}{\alpha_{max}} & (a) \\
-1 & (b)
\end{cases}
, cost_{i,t} = \frac{|\delta_{i,t} - \delta_{i, t-1}|}{z_\delta}
\end{equation}
where $M_i$ is a set of targets selected for the sensor $i$ according to $g_i$;
$cost_{i,t}$ is the power consumption, measured by the normalized moved angle $\frac{|\delta_{i,t} - \delta_{i, t-1}|}{z_\delta}$; $\delta_i$ represents the absolute orientation of sensor $i$, the cost weight $\beta$ is 0.01 and $z_\delta$ is the rotation angle, that is $5^{\circ}$ in our setting.
% \begin{equation}
% \label{eq:tracking_reward}
% r_{i, j, t} = 
% \begin{cases}
% 1 - \frac{|\alpha_{ij, t}|}{\alpha_{max}} & (a) \\
% -1 & (b)
% \end{cases}
% \end{equation}

%Ideally, the executor should adapt to any combination of targets. However, it is difficult to train a neural network to map a noisy observation and the goal vector to goal-conditioned state representation. 
\textbf{Goal-conditioned filter} is introduced to directly remove the unrelated relations based on the assigned goal firstly. With such a clean input, the executor will not be distracted by the irrelevant targets anymore.
For example, if $g_{i}$ is $[1,0,1]$ and $o_{i}$ is $[o_{i,1}, o_{i,2}, o_{i,3}]$, then $\hat{o_i}=filter(o_{i},g_{i}) = [o_{i,1}, o_{i,3}]$. 
In other words, the target-oriented goal can be seen as a kind of hard attention map, forcing the executor only to pay attention to the selected targets.

The network architectures of state encoder, actor and critic are detailed in Appendix~\ref{app5}. The action $a_{i,t}$ is the primitive action and the value $v^L_{i,t}$ estimates the coverage quality of the assigned targets of the sensor $i$. All the executors share the same network parameters.

\vspace{-0.2cm}
\subsection{Training Strategy}
Similar to most hierarchical RL methods, we adopt the two-step training strategy for stability.
It is because that a stochastic executor will lead to a poor team reward, which will bring additional difficulty for learning coordination. At the same time, the coordinator would generate a lot of meaningless goals, e.g. selecting two targets that are far away from each other, which will make the executor confused and waste time on exploration.
Instead, the two-step training can prevent the learning of coordinator/executor from the disturbance of the other.
% prevents the noise generated by other level policy. when training the executor and the coordinator from affecting each other's training.

As for the training of the executor, a goal generation strategy is introduced for training the executor without coordinator. 
% Thus, we design a goal generation method, considering the realistic distribution, e.g., the further targets, the lower probability to be selected. The details of the generation method is described in the Appendix. 1. 
% The pseudo goal generation is introduced for training the executor without coordinator. 
Every $k=10$ time steps, we generate the goal, according to the distance between targets and sensors.
To be specific, the targets, whose distances to sensor $i$ are less than the maximum coverage distance ($\rho_{ij,t}<\rho_{max}$), will be selected as the goal $g_{i,t}$ for sensor $i$.
% if a target $j$ is within the coverage range of sensor $i$, where $\rho_{ij,t}<\rho_{max} \& |\alpha_{ij,t}|<|\alpha_{max}|$, it will be selected for sensor $i$.
Although such strategy mixes some improper targets in the $g_{i,t}$, this will induce a more robust tracking policy for the executor.
% Thus, it is feasible to directly combine the learned executor and the coordinator trained in the next stage while testing.
With the generated goal, We score the coverage quality of the assigned targets for each executor as the individual reward, refer to Eq. \ref{eq:tracking_reward}.
%Under the goal, the local reward is goal-conditioned to score the tracking quality of $g_i$, shown in Eq. \ref{eq:local_reward}. 
Then, the policy can be easily optimized by the off-the-shelf RL method, e.g. A3C~\cite{mnih2016asynchronous}.
% Notably, a well-performed executor will significantly ease the process of learning coordination.

After that, we train the coordinator cooperating with well-performed executors. 
% The interaction between the coordination and executors is periodically. 
While learning, the coordinator updates the observation $\vec{o}$ and goal $\vec{g}$ every $k$ steps.
During the interval, executor $i$ will take primitive actions $a_i$ step-by-step directed by $g_i$.
% executors will continuously take primitive actions from the time step $t$ to $t+ k$, once $\vec{g}$ is given at the time step $t$. 
We fix $k=10$ in the experiments, while learning an adaptive termination (dynamic $k$) is our future work.
The policy is also optimized by A3C~\cite{mnih2016asynchronous}.
We notice that directly applying the executor learned in the previous step will lead to the large decrease of the frame rate (only $25$ FPS), which causes the training of the coordinator time-consuming. 
As an alternative, we build a scripted executor to perform low-level tasks to speed up the training process.
The scripted executor could access the internal state for designing a simple yet effective programmed strategy, detailed in Appendix~\ref{app2}.
Then, the frame rate for the coordinator increases to 75 FPS.
Notably, while testing, we use the learned executor to replace the scripted executor, since the internal state is unavailable in real-world scenarios.
% Since the inference of the trained executor is time-consuming, 
% we adopt a rule-based executor for the training of the coordinator.
% With all of the above, we can train the HMCM from bottom to top.

\vspace{-0.3cm}
\section{Experiments}
First, we build a numerical simulator to imitate the target coverage problem in real-world DSNs.
Second, we evaluate HiT-MAC in the simulator, comparing with three state-of-the-art MARL approaches (MADDPG\cite{lowe2017multi}, SQDDPG\cite{2019arXiv190705707W} and COMA\cite{foerster2018counterfactual}) and one heuristic centralized method (Integer Linear Programming, ILP) for this problem. 
%At the same time, we formulate the problem as an Integer Linear Programming (ILP) problem and solve it by CBC\footnote{https://github.com/coin-or/Cbc} (Coin-or branch and cut) optimizer.
% The complete ILP formulation and other details are described in the appendix. 
We also conduct an ablation study to validate the contribution of the attention module and AMC in the coordinator.
Furthermore, we evaluate the generalization of our method in environments with different numbers of targets and sensors.
The code is available at \url{https://github.com/XuJing1022/HiT-MAC} and the implementation details are in Appendix~\ref{app5}.

\vspace{-0.3cm}
\subsection{Environments}
\label{env}

\begin{wrapfigure}{r}{0.5\linewidth}
\vspace{-1.2cm}
  \centering
  \includegraphics[width=\linewidth]{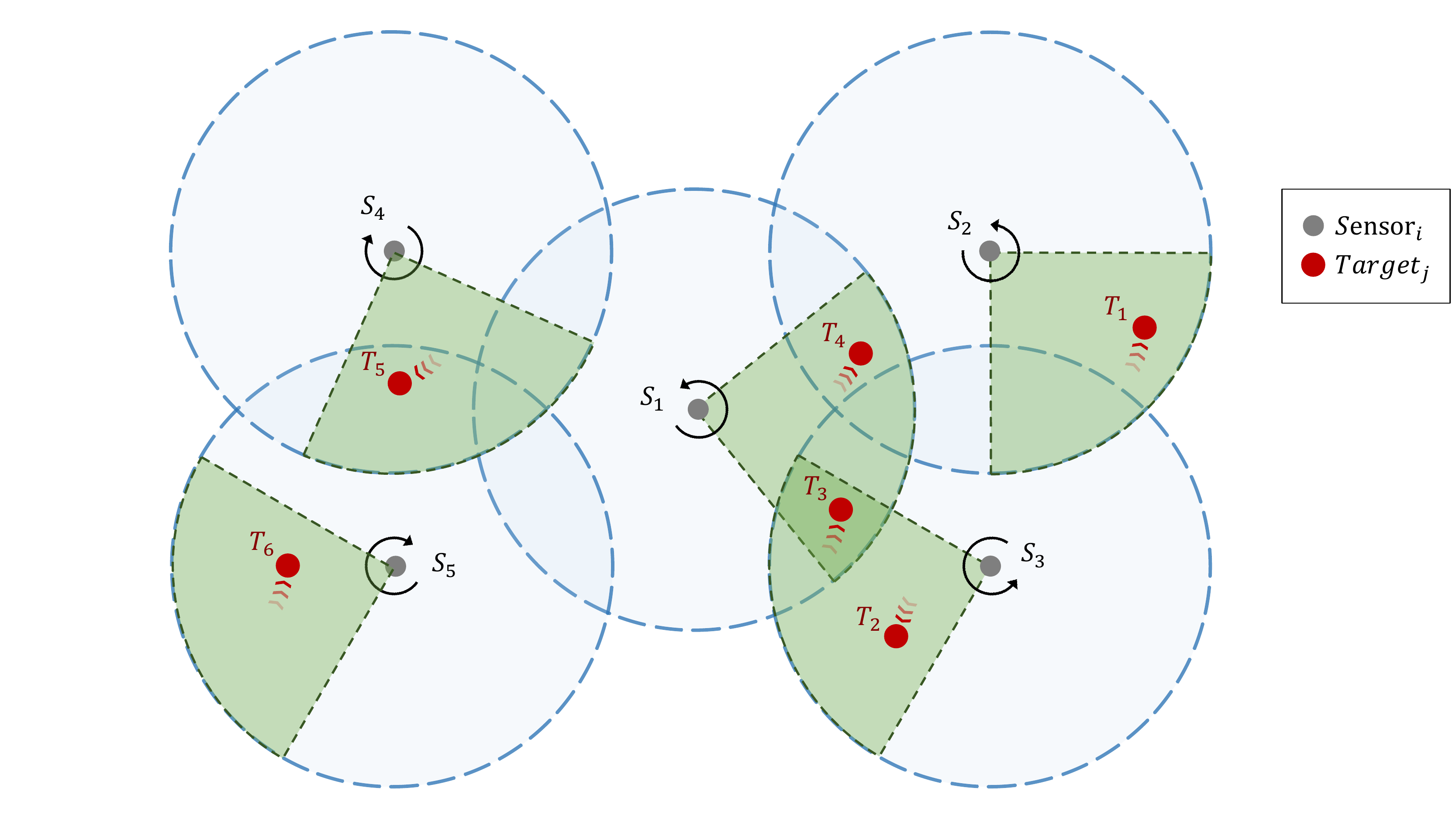}
  \vspace{-0.5cm}
  \caption{An example of the 2D environment.}
  \label{DSN}
\vspace{-0.5cm}
\end{wrapfigure}

To imitate the real-world environment, we build a numerical environment to simulate the target coverage problem in DSNs. 
At the beginning of each episode, the $n$ sensors are randomly deployed. 
Meanwhile, the $m$ targets spawn in arbitrary places and walk with random velocity and trajectories. 
% The trajectories are randomized by randomly sample goal point as the destination for each target
% To diversify the trajectories, we randomly sample goal point as the destination for each target. Once the target arrives, a next new point will be set to arrive.

\textbf{Observation Space.}
In every time step, the observation $o_{i}$ is packed by the sensor-target relations, i.e. $o_i=(o_{i,1}, o_{i,2},..., o_{i,m})$.
$o_{i,j}=(i, j, \rho_{ij}, \alpha_{ij})$ describes the spatial relation between sensor $i$ and target $j$ in a polar coordinate system (the sensor $i$ is at the origin). 
To be specific, $i$, $j$ are the ID of the sensor and target separately; $\rho_{ij}$ and $\alpha_{ij}$ are the absolute distance and relative angle from $i$ to $j$.
% $o_{i,j}$ is a 4-dimensional vector, which consists of
% \begin{itemize}
%     \item $f_1$: sensor ID $i$,
%     \item $f_2$: target ID $j$,
%     \item $f_3$: angle error between $i$ and $j$
%     \item $f_4$: distance between $i$ and $j$
% \end{itemize}
% In such state representation, IDs are for distinguishing. $f_3$ and $f_4$ describe the visibility of the target $j$. $f_5$ and $f_6$ are position relation between the agent $i$ and the target $j$.
For the coordinator in HiT-MAC, it takes $\vec{o} = (o_{1}, ..., o_{n})$ as the joint observation.
% And the observations are filtered by $\Vec{g}$ for executors as the inputs. The executors, namely sensors, make decisions among the primitive actions according to the filtered observations.

\textbf{Action Space.} 
The primitive action space is discrete with three actions $Turn Right$, $Turn Left$ and $Stay$. 
Quantitatively, the $Turn Right/Trun Left$ will incrementally adjust the sensor's absolute orientation $\delta_i$ in $5$ degree, i.e.,$Right$:$\delta_{i,t+1}+=5$, $Left$:$\delta_{i,t+1}-=5$.
% to change views of sensors. The corresponding quantitative representations are \{{(5), (-5), (0)}\}, which are direction actions deciding the rotating angles of sensors.
For the coordinator in HiT-MAC, the goal map $\Vec{g}$ is a $n \times m$ binary matrix, where $g_{i,j}$ represents whether the target $j$ is selected for the sensor $i$ (0: No, 1: Yes).
Each row corresponds to the assignment for each sensor one by one.

\vspace{-0.3cm}
\subsection{Evaluation Metric}

We evaluate the performance of different methods on two metrics: coverage rate and average gain.
\textbf{Coverage rate (CR)} is the primary metric, measuring the percentage of the covered targets among all the targets, shown in Eq. \ref{eq:global_reward}; 
\textbf{Average gain (AG)} is an auxiliary metric to measure the efficiency in power consumption. It counts how much CR gains each rotation brings, i.e. $CR / \widehat{cost}$, where $\widehat{cost} = \frac{1}{Tn}\sum_{t=1}^{T}\sum_{i=1}^{n} cost_{i,t}$, and $cost_{i,t}$ is previously introduced in Eq.~\ref{eq:tracking_reward}.
% where $\hat{N}_{rotation}=N^{tot}_{rotation}/n$ and $N^{tot}_{rotation}$ is the times that the actions $TurnRight$/$TurnLeft$ are taken by each sensor per episode.

For good performance, we expect both metrics to be high.
In practice, we consider the CR in primary. Only when methods achieve comparable CR, AG is meaningful.
% A well-worked DSN system guided by the optimal coordination should cover targets as much as possible with a higher average gain, to guarantee that all primitive actions taken by sensors are indispensable and effective and the cost caused by rotation is worth as much as possible. 
For mitigating the bias caused by randomness of training and evaluation, we count the results and draw conclusions after running training for 3 times and evaluation for 20 episodes.

\vspace{-0.3cm}
\subsection{Baselines}
We employ MADDPG\cite{lowe2017multi}, SQDDPG\cite{2019arXiv190705707W} and COMA\cite{foerster2018counterfactual}, three state-of-the-art MARL methods as baselines. 
% The codes are from \url{https://github.com/hsvgbkhgbv/SQDDPG}.
They all are trained with a centralized critic and executed in a decentralized manner. 
But their critics are built in different ways for credit assignment, e.g., SQDDPG\cite{2019arXiv190705707W} aims at estimating the shapley Q-value for each agent. 
As for the target coverage problem in DSNs, one heuristic method is to formulate the problem as an integer linear programming (ILP) problem and globally optimize it at each step. See Appendix~\ref{app3}\&\ref{app4} for more details.

\begin{figure*}[t]
\centering
\vspace{-0.5cm}
\subfigure[]{
    \label{method_compare}
    \includegraphics[width=0.47\linewidth]{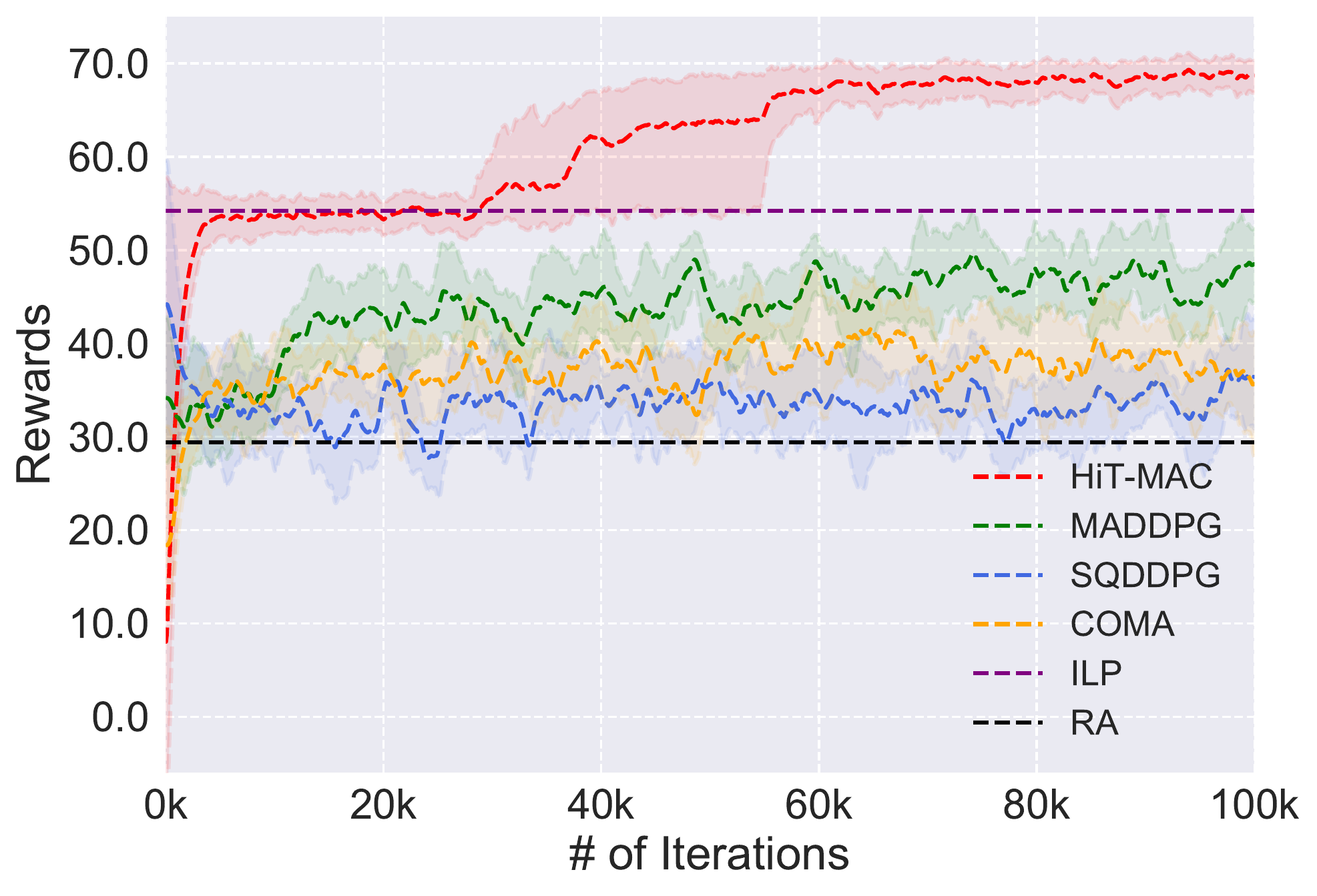}
    }
    \subfigure[]{
    \label{4v5_ablation}
    \includegraphics[width=0.47\linewidth]{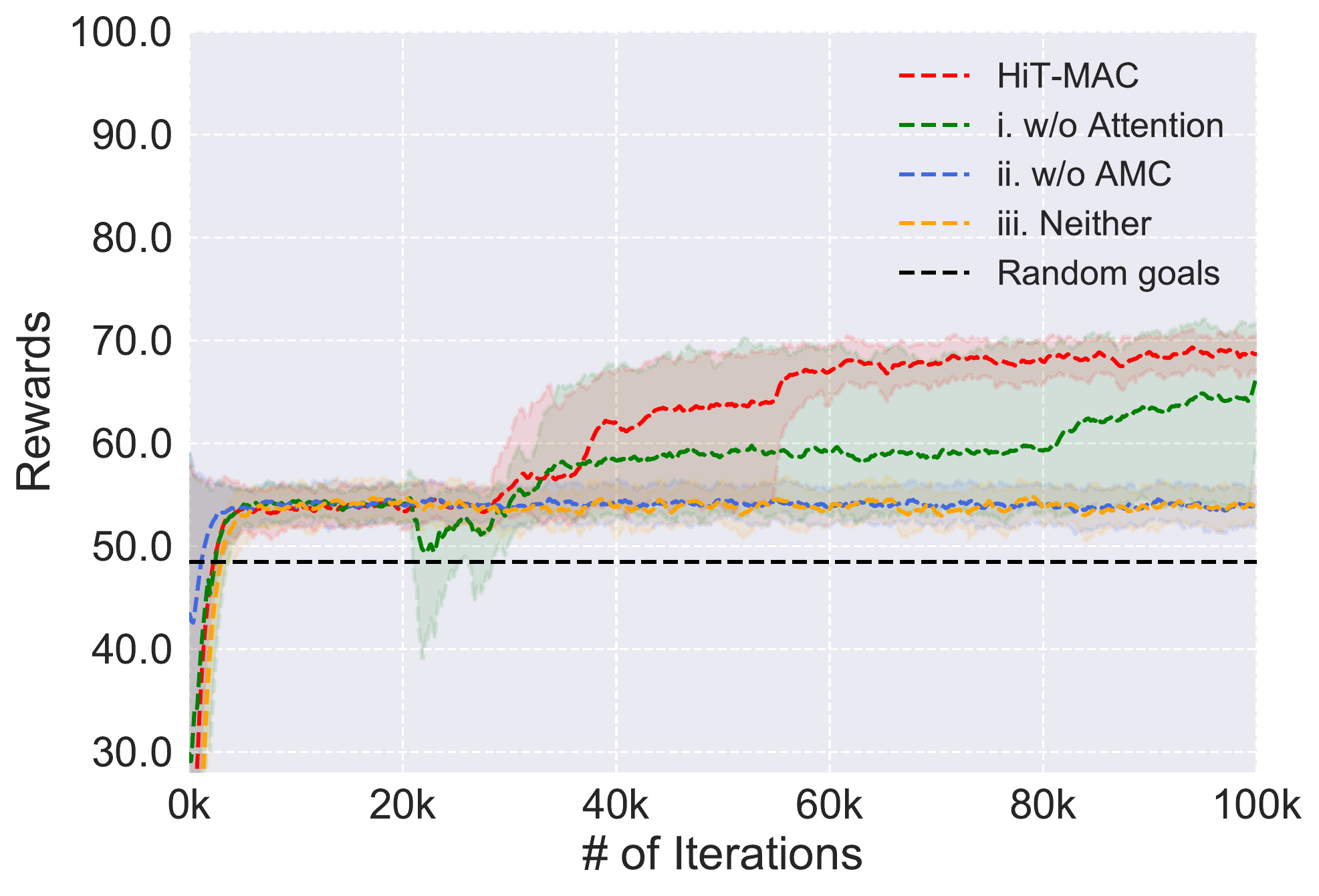}
    }
\vspace{-0.3cm}
  \caption{
  The learning curve of all learning-based methods. They all are trained in environment with 4 sensors and 5 targets.
  (a) comparing ours with baselines; (b) comparing ours with its ablations.}
  \vspace{-0.5cm}
\end{figure*}
% \begin{wrapfigure}{r}{0.4\linewidth}
% \centering
% \subfigure[]{
%     \label{method_compare}
%     \vspace{-0.6cm}
%     \includegraphics[width=\linewidth]{figures/baseline_4v5.pdf}
%     }
%     \subfigure[]{
%     \label{4v5_ablation}
%     \includegraphics[width=\linewidth]{figures/4v5_ablation.pdf}
%     }
%   \vspace{-0.4cm}
%   \caption{
%   The learning curve of all the MARL methods. They all are trained and tested in environments with 4 sensors and 5 targets.
%   (a) comparing ours with baselines. (b) comparing ours with its ablations}
% \end{wrapfigure}

\begin{wraptable}{r}{0.45\linewidth}
\vspace{-0.7cm}
  \caption{Comparative results of different methods (n=4\&m=5). }
  \label{different-methods}
  \centering
  \resizebox{\linewidth}{!}{
  \begin{tabular}{lcc}
    \toprule
    Methods    & Coverage rate( \% ) $\uparrow$    & Average gain $\uparrow$ \\
    \midrule
    MADDPG    & 45.56$\pm$9.45    & 1.38\\
    SQDDPG    & 36.67$\pm$9.04   & 2.73  \\
    COMA      & 35.37$\pm$8.41    & 2.49\\
    ILP       & 54.18$\pm$12.32    & \textbf{3.87}\\
    \hline
    HiT-MAC(Ours)    & \textbf{72.17}$\pm$\textbf{5.58}   & 1.46   \\
    \bottomrule
  \end{tabular}
  }
  \vspace{-0.3cm}
\end{wraptable}

% \begin{wrapfigure}{r}{0.4\linewidth}
%   \vspace{-1.0cm}
%   \centering
%   \subfigure[]{
%     \label{method_compare}
%     \vspace{-0.6cm}
%     \includegraphics[width=\linewidth]{figures/baseline_4v5.pdf}
%     }
%     \subfigure[]{
%     \label{4v5_ablation}
%     \includegraphics[width=\linewidth]{figures/4v5_ablation.pdf}
%     }
%   \vspace{-0.4cm}
%   \caption{
%   The learning curve of all the MARL methods. They all are trained and tested in environments with 4 sensors and 5 targets.
%   (a) comparing ours with baselines. (b) comparing ours with its ablations}
% \end{wrapfigure}

\subsection{Results}
\vspace{-0.3cm}
\textbf{Compare with Baselines.}
As Fig.~\ref{method_compare} shows, our method achieves the highest global reward in the setting with 4 sensors and 5 targets.
We also draw the performance of ILP and random agents in Fig.~\ref{method_compare} as reference.
We can see that state-of-the-art MARL methods work poorly in this setting.
None of them could exceed the ILP. 
Typically, the improvement of SQDDPG is marginal to agents with random actions.
This suggests that it is difficult to directly estimate the marginal contribution of each agent in this problem.
% This indicates that it is not a good option to use the fully decentralized MARL methods for this problem.
Instead, HiT-MAC surpasses all the baselines after $35k$ of iterations, and reaches a stable performance of $\sim70$ at the end.
As for the quantitative results during evaluation in Tab.~\ref{different-methods}, HiT-MAC consists of the trained coordinator and trained executors and significantly outperforms the baselines in CR.
ILP gets the highest AG, as it globally optimizes the joint policy step-by-step. 
COMA and SQDDPG also get higher AG than ours, but in fact, they only learn to take no-operation to wait for the targets run into its coverage range. As a result, their CRs are lower than others.

\begin{wrapfigure}{r}{0.35\linewidth}
\vspace{-0.5cm}
  \centering
  \subfigure[]{
    \label{generalization.4vm}
    \includegraphics[width=\linewidth]{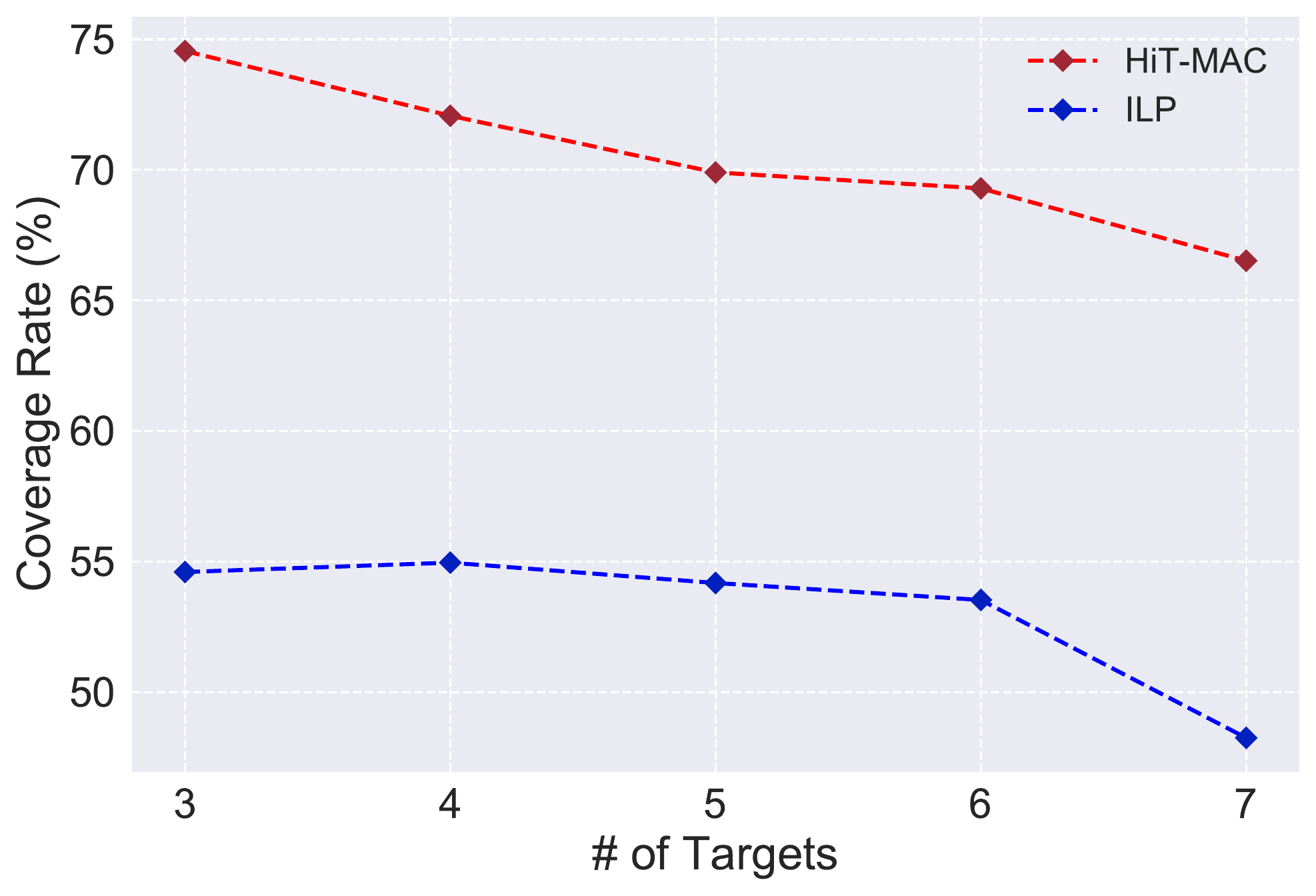}
    }
    % \vspace{-0.7cm}
    \subfigure[]{
    \label{generalization.nv5}
    \includegraphics[width=\linewidth]{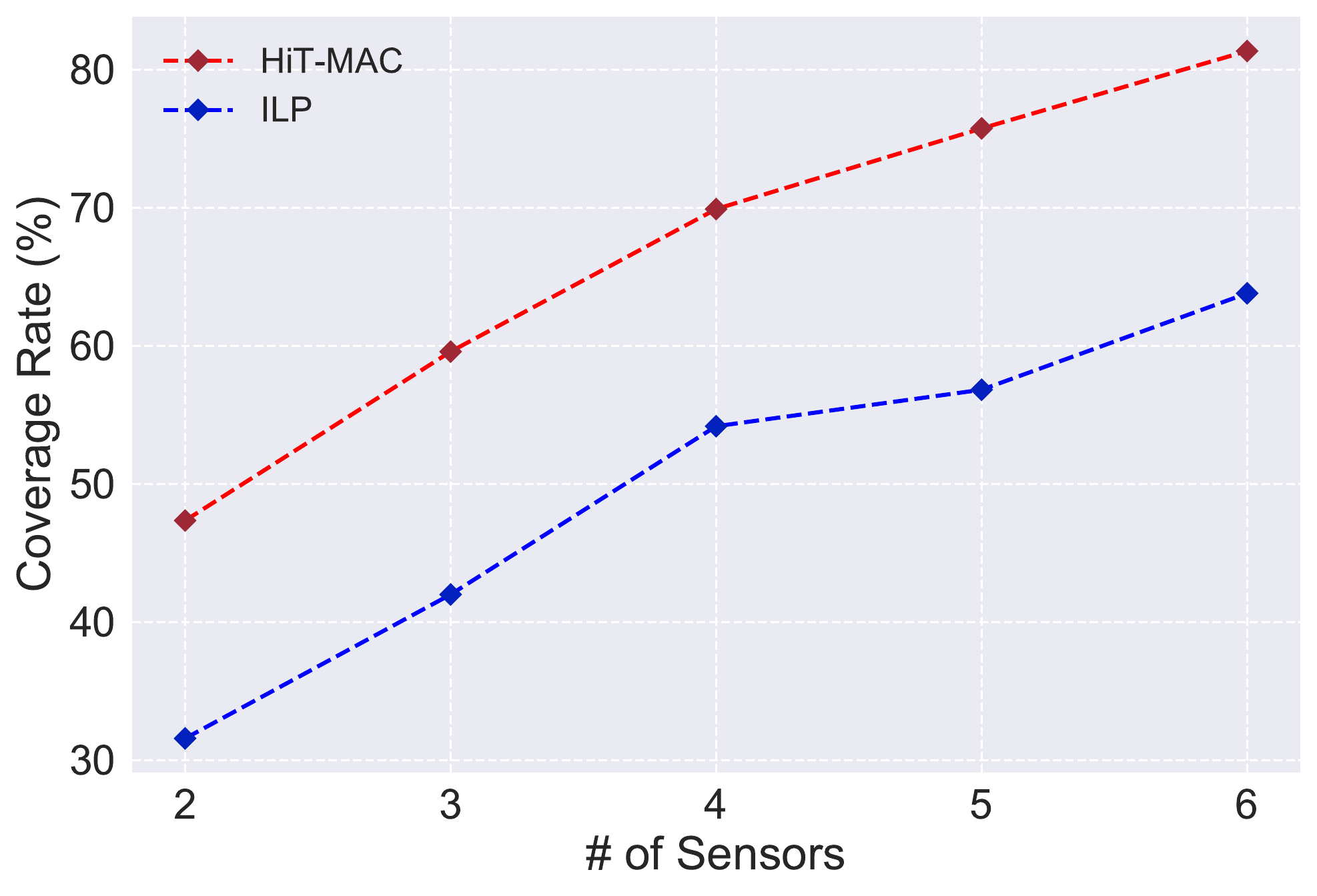}
    }
    \vspace{-0.3cm}
  \caption{Analyzing the generalization of HiT-MAC to the different number of sensors $n$ and targets $m$. (a) $n=4$, $m$ is from $3$ to $7$; (b) $m=5$, $n$ is from $2$ to $6$.}
\vspace{-0.3cm}
\end{wrapfigure}

\textbf{Ablation Study.}
We consider ablations of our method that help us understand the impact of attention framework and global value estimation by AMC shown in the Fig. \ref{4v5_ablation}. We compare our method with (\rmnum{1}) the one without attention encoder; (\rmnum{2}) the one without AMC; (\rmnum{3}) the one without attention encoder nor AMC in  $n=4\&m=5$. For (\rmnum{1}) and (\rmnum{3}), we use bi-directional Gated Recurrent Unit (BiGRU~\cite{schuster1997bidirectional,chung2014empirical}) to replace the attention module. As for the critic input, we use context feature $C_t$ for (\rmnum{2}) and the hidden state of BiGRU for (\rmnum{3}).
From the learning curve, we can see that the ablations that without AMC (\rmnum{2}, \rmnum{3}) stuck in a locally optimal policy.
Their rewards are close to the random policy, which randomly samples the targets as the goal for executors.
Instead, the performance of (\rmnum{1}) and ours could be further improved after $35k$ of iterations. 
This evidence demonstrates that the introduced AMC method is capable of effectively guiding the coordinator to learn a high-quality target assignment.
Compared with (\rmnum{1}), our method with attention-based encoder converges faster and more stably.
And the variance of the training curve in (\rmnum{1}) is larger than ours, though the one without attention-based encoder can also converge to a high score sometimes.
So, we think that the attention-based encoder is more suitable for the coordinator, rather than RNN.
It is because that attention mechanism can aggregate the features without any assumption about the sequence order, rather than following a specific order to encode the data.

\textbf{Generalization.} 
We analyze the generalization of our method to the different number of sensors $n$ and targets $m$.
% We test ours and ILP in some unseen environment settings by changing the complexity of the environment from two aspects, i.e. the number of the sensors $N$ and the number of the targets $M$.
While testing, we adjust the number of targets and sensors in the environment, respectively, and report the results under each setting for better analysis. 
For example, in Fig.~\ref{generalization.4vm}, we can see the trend of performance with the change of the target number from 3 to 7 in the 4 sensors case.
In the same way, we also demonstrate the trend of performance in the cases of varying sensor numbers ($n=2,3,4,5,6 \& m=5$), shown as Fig.~\ref{generalization.nv5}.
Note that our model is only trained in a fixed-number environment ($n=4 \& m=5$). 
We report the rewards of ILP as reference, as its performance does not depend on the training environment and has stable generalization in different settings. 
Since the score of ILP is already lower than ours, we compare the changing of the score to ensure as much fairness as possible. 
% In Fig. \ref{generalization.4vm} and \ref{generalization.nv5}, ours is the model trained in the N=4 \& M=5.
In Fig. \ref{generalization.4vm}, the performance of ours increases stably with the decrease of $m$ from 5 to 3, while the reward of ILP increases lightly. 
With the increase of $m$ from 5 to 7, ours decreases slower than ILP.
In Fig. \ref{generalization.nv5}, our score increases more stably than ILP when $n$ increases from 4 to 6. 
Those can be concluded that HiT-MAC is scalable and of a good generalization in environments with different numbers of sensors and targets.

\vspace{-0.3cm}
\section{Related Work}
\vspace{-0.3cm}
\textbf{Coverage Problem} is a crucial issue of directional sensor networks~\cite{ma2007some}. The available studies about coverage problem can be categorized into four main types~\cite{guvensan2011coverage}: target-based coverage, area-based coverage, sensor deployment, and minimizing energy consumption. And a set of heuristic algorithm~\cite{han2010greedy, li2018heading, zhang2016local} has been proposed to find a nearly-optimal solution under a specific setting, as most of the problems are proved NP-hard. 
Recently, with the advances of machine learning, Mohamadi, et al. adopt learning-based methods~\cite{mohamadi2015new, mohamadi2014learning} for maximizing network lifetime in wireless sensor networks.
However, all the algorithm are designed for a specific setting/goals.
In this work, we focus on finding a non-trivial learning approach for the target-based coverage problem. We formulate the coverage problem as a multi-agent cooperate game, and employ the modern multi-agent reinforcement learning to solve the game. 
%The change of objective function can be simply reflected in the reward function. 

\textbf{Cooperative Mutli-Agent Reinforcement Learning(MARL)} addresses the sequential decision-making problem of multiple autonomous agents that operate in a common environment, each of which aims to collaboratively optimize a common long-term return~\cite{bu2008comprehensive}. With the recent development of deep neural network for function approximation, many prominent multi-agent sequential decision-making problems are addressed by MARL, e.g. playing real-time strategy games~\cite{vinyals2019alphastar}, traffic light control~\cite{chu2019multi}, swarm system\cite{hung2016q}, common-pool resolurce appropriation~\cite{perolat2017multi}, sequential social dilemmas~\cite{leibo2017multi}, etc.
In Cooperative MARL, it is notoriously difficult to give each agent an accurate contribution under a shared reward. This phenomenon limits the further application of MARL in more difficult problems, referred as credit assignment.
This motivates the study on the local reward approach, which aims at decomposing the global reward to agents according to their contributions.
~\cite{nguyen2018credit, foerster2018counterfactual} modeled the
contributions inspired by the reward difference. Based on shapley value~\cite{shapley1953stochastic}, shapley Q-value~\cite{2019arXiv190705707W} is proposed to approximate a local reward, which considers all possible orders of agents to form a grand coalition. In this paper, we learn the critic in coordinator by approximating the marginal contribution of each sensor-target assignment for effective learning the coordination policy.

% 实现sensor之间协作的方法有centralized，decentralized。centralized有什么优缺点，decentralized有什么优缺点。
% To build such a multi-agent system, there are mainly two classes of approaches available: centralization and decentralization.
% In a classic centralized system, there is a central agent monitoring the global state and controlling all sensors’ motion at each time step. 
% In this case, the central policy can be viewed as a single agent and easily optimized by the off-the-shelf reinforcement learning methods.
% However, it is hard to satisfy the requirement in server-sensors communication, which should keep high bandwidth and low delay for real-time interaction. This is challenging for sensor networks, especially with the increasing number of sensors.
% % policy can be optimzied by a single-agnet RL method, but high-cost in communication and weak in scability. 
% In a decentralized system, each sensor is controlled by its own agent independently, which observe the environment by themselves and exchange their observations with others by communication.
% Such a decentralized system could run in a larger scale multi-agent system, be robust to the change of scale, and save the cost in communication. But in most cases, the distributed policy is unstable and difficult to optimize jointly, as they usually affect others.

\vspace{-0.2cm}
\section{Conclusion and Discussion}
\vspace{-0.2cm}
In this work, we study the target coverage problem, which is the main challenge problem in the DSNs. We propose an effective Hierarchical Target-orient Multi-agent Coordination framework (HiT-MAC) to further enhance the coverage performance. In HiT-MAC, we decompose the coverage problem into two subtasks: assigning targets to sensors and tracking assigned targets.
To implement it, we further introduces a bunch of pratical methods, such as AMC for critic, attention mechanism for state encoder.
Empirical results demonstrated that our method can deal with challenging scenes and outperform the state-of-the-art MARL methods.

Although significant improvements have achieved by our methods, there is still a set of drawbacks waiting for addressed. 1) 
% The goal space grows exponentially in the number of sensors and targets $2^{nm}$. 
We need to find a solution to deploy the framework in large scale DSNs ($n>100$), e.g. multi-level hierarchy.  
% To do this, we need to find a solution to extend the two-level hierarchy to a multi-level hierarchy, or cluster the sensors and targets to reduce the observation space.
2) For a practical application, it is necessary to additional consider a more real-world setting, including placing obstacles, using visual observation, and limited communication bandwidth.
3) For the executor, it is required to learn an adaptive termination, rather than the fixed k-step execution.
Furthermore, it is also an interesting future direction to apply our method on other target-oriented multi-agent problems, where agents are focused on optimizing some relations to a group of targets, e.g. collaborative object searching\cite{lowe2017multi}, active object tracking~\cite{zhong2019ad}.
% Those are both interesting avenues for our future work.

\section*{Broader Impact}
The target coverage problem is common in Directional Sensor Networks. This problem widely exists in a lot of real-world applications.
For example, those who control the cameras to capture the sports match videos may benefit from our work, because our framework provides an automatic control solution to free them from heavy and redundancy labor. 
Surveillance camera networks may also benefit from this research. 
But there is also the risk of being misused in the military field, e.g., using directional radar to monitor missiles or aircraft.
The framework may also inspire the RL community, for solving the target-oriented tasks, e.g. collaborative navigation, Predator-prey.
If our method fails, the targets would be all out of views of sensors. 
So, maybe a rule-based alternate plan is needed for unexpected conditions.
We reset the training environment randomly to leverage biases in the data for better generalization.

\begin{ack}
We thank Haifeng Zhang, Wenhan Huang, and Prof. Xiaotie Deng for their helpful discussion in our early work. 
This work was supported by MOST-2018AAA0102004, NSFC-61625201, the NSFC/DFG Collaborative Research Centre SFB/TRR169 "Crossmodal Learning" II, Qualcomm University Research Grant, Tencent AI Lab RhinoBird Focused Research Program (JR201913).
\end{ack}

% \section*{References}
{\small
\bibliographystyle{ieeetr}
\bibliography{ref}
}

\clearpage
\appendix
% \appendixautorefname
% \setcounter{section}{0}

\begin{center}
\LARGE{\textbf{Appendix}}
\end{center}

\section{Goal generation for executor training}
The pseudo goal generation is introduced for training the executor without coordinator. 
Every 10 time steps, we generate the goal, according to the distance between targets and sensors. 
To be specific, the targets, whose distances to sensor $i$ are less than the maximum coverage distance ($\rho_{ij,t}<\rho_{max}$), will be selected as the goal $\Vec{g_{i,t}}$ for sensor $i$.
Although the generated goals are not aligned to the policy of the ideal coordination,
some improper targets mixed in the $\Vec{g_{i,t}}$ can lead the executor to learn a more robust policy to track the assigned targets.
Then, such a robust executor may adapt to the coordinator trained in the next stage better.

\section{Programmed strategy for the executors}
\label{app2}
we build a scripted policy to perform low-level tasks while training the coordinator, since directly applying the learned executor will cause the frame rate to drop severely (only $25$ FPS).
The scripted policy is allowed to access the grounded state, e.g. the absolute position $(x_j, y_j)$ of each targets and the pose of the sensor.
% the absolute position $(x_i, y_i)$ and the orientation $rot_i$ in two-dimensional world coordinate system. 
Intuitively, one feasible solution to track a set of targets is to make the sensor pointing to the cluster center of the targets.
% Thus, we use the grounded state to compute the the center of all the assigned targets, and 
Thus, we calculate the center $(x^M_{mean}, y^M_{mean})$ of the assigned targets $M_i$ for sensor $i$, while the script policy is to take primitive actions to minimize the relative angle to the center.
Given the pose of sensor $i$ is $(x_i, y_i, \alpha_i)$ where $(x_i, y_i)$ is the position and $\alpha_i$ is the orientation, the relative angle error $\beta_i$ is calculated as 
\begin{equation}
    arctan(\frac{y^M_{mean}-y_i}{x^M_{mean}-x_i})*\frac{180}{\pi}-\alpha_i
\end{equation}
And, the taken action $a_{i,t} = clip(\beta_i//z_\delta, -1, 1)$. Here $z_\delta=5$, because the action is deterministic and the rotation unit is 5 degree.
With this trick, the frame rate for training of the coordinator increases to 75 FPS.
Note that it is not the optimal policy for the executor, it will fail when two targets are far.

\section{Integer Linear Programming}
\label{app3}
\begin{wrapfigure}{r}{0.4\linewidth}
  \centering
  \includegraphics[width=0.8\linewidth]{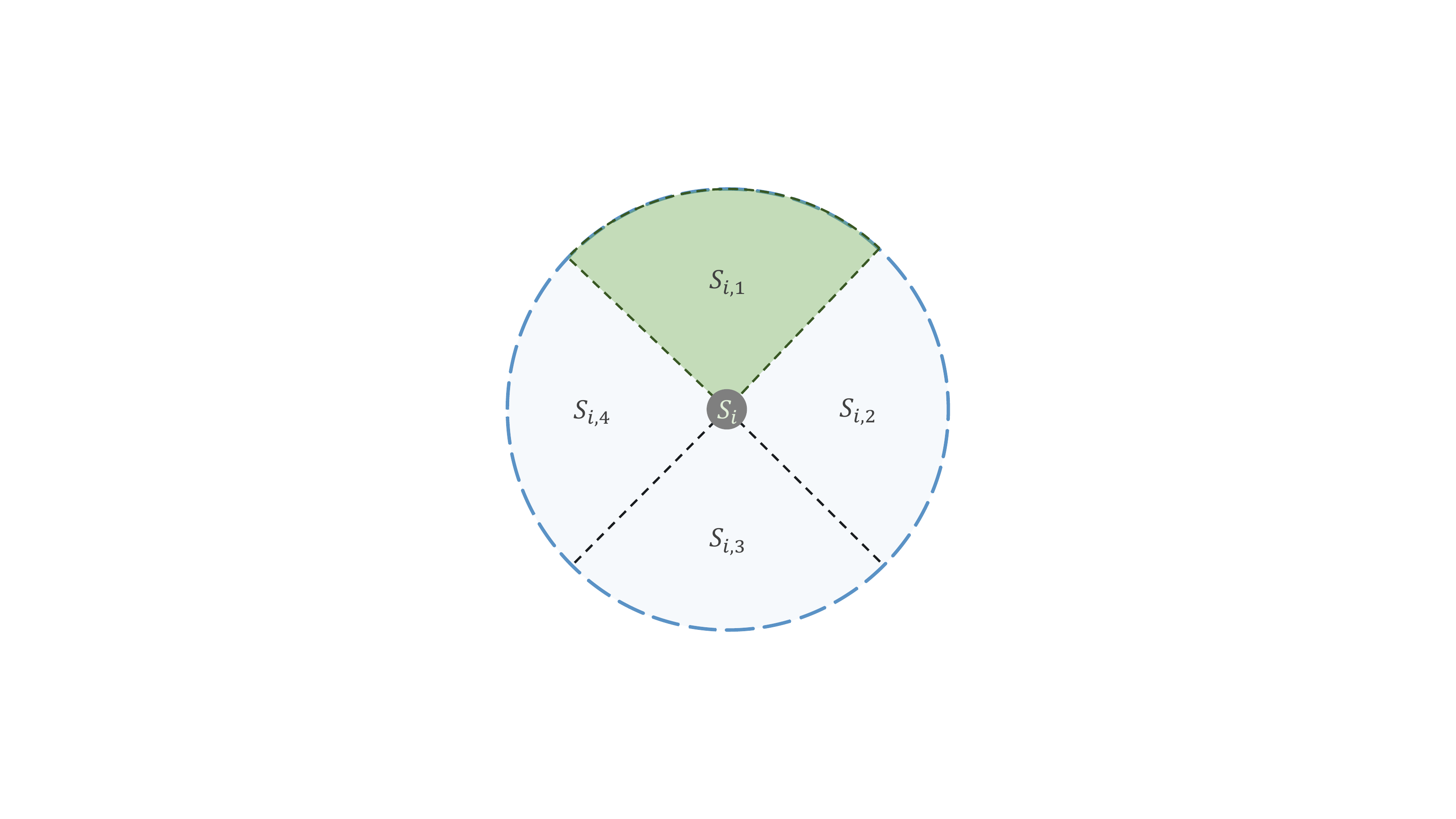}
  \caption{Optional direction partition}
  \label{nodes}
\end{wrapfigure}

As one of the baselines, we formulate the problem as an Integer Linear Programming (ILP) problem and solve it by CBC\footnote{https://github.com/coin-or/Cbc} (Coin-or branch and cut) optimizer. The inputs are position, rotation, sensing range of sensors and position of targets, while the outputs are the primitive actions taken by every sensor, i.e. $Turn Right$, $Turn Left$ or $Stay$.

The notations used here are defined as follows.
\begin{itemize}
    \item $N$: the number of the sensors
    \item $P$: the number of optional directions for a sensor
    \item $S$: the set of sensors
    \item $s_i$: the i-th sensor
    \item $s_{i,j}$: the j-th direction of the i-th sensor
    \item $R_{i,j}$: the monitor region of $s_{i,j}$
    \item $M$: the number of the targets
    \item $T$: the set of targets
    \item $t_k$: the k-th target
\end{itemize}

Consider a randomly deployed directed sensor network, there are $N$ directed sensors $S$ to monitor the targets $T$, and each sensor has $P$ optional directions. 
First of all, we set P as 4 shown in Fig\ref{nodes}, because the visual angle is $90^{\circ}$. $S_{i,1}$ is always the current direction of the sensor $i$. The other optional directions are clockwise numbered as $S_{i,j}$. And the primitive actions depends on the selected directions solved from ILP.

The variables in the ILP are described as follows: the binary variable $y_k$ is 1 if and only if the target $t_k$ is covered by an arbitrarily sensor, otherwise it is 0; the binary variable $x_{i,j}$ takes the value 1 if and only if the i-th sensor is in the j-th direction, otherwise it takes 0.

For each directed sensor, define a matrix whose element $a^i_{j,k}$ indicates whether the target $t_k$ is in the $s_{i,j}$ sensing area:

\begin{equation}
a^i_{ j, k} = 
\begin{cases}
1, & t_k \in R_{i,j} \\
0, & otherwise
\end{cases}
     \forall j=1,\cdots,P; \forall k=1,\cdots,M
\end{equation}

Define a non-negative integer $\delta_k$ as the number of sensors that cover the target $t_k$:

\begin{equation}
    \delta_k=\sum^N_{i=1} \sum^P_{j=1} a^i_{ j, k} x_{i,j} ,  \forall k=1,\cdots,M
\end{equation}

So, the target coverage problem can be formulated as Integer Linear Problem as follows:

\begin{equation*}
\centering
\begin{split}
&\max \,\, z=\sum^M_{k=1} y_k\\
&s.t.\quad  \left\{\begin{array}{lc}
\delta_k/N\leq y_k \leq \delta_k, &\forall k=1,\cdots,M\\
\sum^P_{j=1} x_{i,j}\leq 1,&\forall i=1,\cdots,N\\
y_k=0 \, or \, 1, & \forall k=1,\cdots, M\\
x_{i,j} = 0 \, or \, 1, & \forall i=1,\cdots,N; \forall j=1,\cdots P \end{array}\right.
\end{split}
\end{equation*}

The objective is to maximize the number of covered targets. The first constraint denotes whether $t_k$ is covered. The second one denotes that a sensor can only work in one of the optional directions at the same time.

After formulation, we can solve the target coverage problem as an ILP problem with CBC optimizer. 
Then, the primitive actions for all the sensors can be derived from the results of ILP shown as Tab. \ref{derivation}. 

\begin{table}[h]
  \caption{Derivation rules}
  \label{derivation}
  \centering
  \begin{center}
  \begin{tabular}{cc}
    \toprule
    $x_{i,j}$    &  primitive action \\
    \midrule
      $x_{i,1}$=1      & $Stay$  \\
      $x_{i,2}$=1      & $TurnRight$  \\
      $x_{i,3}$=1      & $TurnRight$ or $TurnLeft$  \\
      $x_{i,4}$=1      & $TurnLeft$  \\
    \bottomrule
  \end{tabular}
  \end{center}
\end{table}

%整数规划是对静态分配更好的，像我们这时序上变动；对视野半径内的分配比较容易，分配视野范围外的target难, 没有对future的prediction.
%整数规划如果用覆盖率为目标函数 那么只有当target离开当前视野 才能采取动作
%但是如果用目标的角度偏差作为要优化的目标函数，对于最大化覆盖率的任务本身，也不太合适.约束的手工设计复杂
In the section 4.3, we can see that the performance of ILP is poor. The reasons are three folds. 
\begin{itemize}
    \item 1) Integer programming is better for static allocation, while targets is preferably within the field of view radius. But, the targets are mobile in our settings. So, when targets are outside the field of view, it needs prediction based on sequential history observations.
    \item 2) The objective function is to maximize the coverage rate, therefore the rotation can only be taken when some target leaves the current field of view, which will lead to lower score. Actually, sensors should be rotated in advance if possible in order to avoid targets loss.
    \item 3) There may be some solutions for the above two problems. For example, handcraft the objective function with intuition, e.g. additionally considering the average relative angle error to the targets. But manually design the constraints will be trivial and difficult.
\end{itemize}

\section{MARL baselines}
\label{app4}
We implement the MARL baselines by employing the codes from \url{https://github.com/hsvgbkhgbv/SQDDPG}.
To be specific, the policy and critic network both are two layers MLP for MADDPG and COMA, as the same as the policy network and hyper network for mixing Q value in the Q-mix.
% The frameworks are similar with those proposed in their papers.
The common hyper-parameters are detailed in Tab. \ref{baseline-hyper}.

\begin{table}[H]
  \caption{Hyper-parameters for baselines}
  \label{baseline-hyper}
  \centering
  \begin{center}
  \begin{tabular}{lcc}
    \toprule
    Hyper-parameters    &   \#   &   Description \\
    \midrule
  hidden units          & 128       & the \# of hidden units for all layers  \\
  training episodes     & 50k       & maximum training episodes  \\
  episode length        & 100       & maximum time steps per episode  \\
  discount factor       & 0.9       & discount factor for rewards, i.e. gamma  \\
  entropy weight        & 0.001      & parameter for entropy regularization\\
  learning rate         & 5e-4    & learning rate for all networks \\
  target update frequency        & 100    & target network updates every \# steps \\
  target update rate        & 0.1       & target network update rate \\
  replay buffer               & 1e4        & the size of replay buffer\\
  batch size      & 64      & the \# of transitions for each update\\
    \bottomrule
  \end{tabular}
  \end{center}
\end{table}

\section{Networks and Hyper-parameters for HiC-MAC}
\label{app5}
As for coordinator, the encoder consists of 2 fully connected(FC) layers and an attention module, the actor consists of one FC layer and the critic based AMC consists of an attention module and one FC layer.
As for executor, the encoder is an attention module, the actor and the critic both consist of one FC layer simply.
The training framework is like A3C, and the hyper-parameters for our method are detailed in Tab. \ref{Hyper-parameters}.

\begin{table}[H]
  \caption{Hyper-parameters for both coordinator and executor.}
  \label{Hyper-parameters}
  \centering
  \begin{center}
  \begin{tabular}{lcc}
    \toprule
    Hyper-parameters    &   \#   &   Description \\
    \midrule
  hidden units          & 128       & the \# of hidden units for all layers  \\
  training episodes     & 50k       & maximum training episodes  \\
  episode length        & 100       & maximum time steps per episode  \\
  discount factor       & 0.9       & discount factor for rewards  \\
  entropy weight        & 0.01      & parameter for entropy regularization\\
  learning rate         & 5e-4    & learning rate for all networks\\
  workers               & 6        & the \# of workers in the A3C framework\\
  update frequency      & 20      & the master network updates every \# steps in A3C\\
    \bottomrule
  \end{tabular}
  \end{center}
\end{table}

\section{Discussion about AMC}
\label{app6}
\subsection{Compare with Shapley Q value in SQDDPG}
\label{app6.1}
The marginal contribution of each coalition in SQDDPG is defined as $\Phi_{i}(\mathcal{C})=Q^{\pi_{\mathcal{C} \cup\{i\}}}\left(s, \mathbf{a}_{\mathcal{C} \cup\{i\}}\right)-Q^{\pi_{\mathcal{C}}}\left(s, \mathbf{a}_{C}\right)$. And they model a function to approximate the marginal contribution directly such that $\hat{\Phi}_{i}\left(s, \mathbf{a}_{\mathcal{C} \cup\{i\}}\right): \mathcal{S} \times \mathcal{A}_{\mathcal{C} \cup\{i\}} \mapsto \mathbb{R}$, where $\mathcal{S}$ is the state space; $\mathcal{C}$ is the ordered coalition that agent $i$ would like to join in; $\mathcal{A}_{\mathcal{C} \cup\{i\}}=\left(\mathcal{A}_{j}\right)_{j \in \mathcal{C} \cup\{i\}}$ and the actions are ordered. 

In SQDDPG, AMC is conducted on $Q(s,a)$ value, which would introduce an extra assumption, i.e. the actions taken in $C$ should be the same as the ones in the coalition $C \cup \{i\}$.
In fact, the actions of every agent in different coalitions are not necessarily the same. 
Here is an example. Given three agents cooperating to complete a task, their optimal joint action is $\mathbf{a}=(a_0,a_1,a_2)$.
If the order of a coalition is (0,2), then $\mathbf{a}_{\mathcal{C} \cup\{1\}} = (a_0, a_2, a_1)$. However, if there are just $agent_0$ and $agent_2$ that cooperate in the environment, the optimal actions $(\hat{a}_0,\hat{a}_2)$ for coalition (0,2) may be different from $(a_0, a_2)$. Actually, the marginal contribution should be $\Phi_{i}(\mathcal{C})=Q^{\pi_{\mathcal{C} \cup\{i\}}}\left(s, (a_0,a_1,a_2)\right)-Q^{\pi_{\mathcal{C}}}\left(s, (\hat{a}_0,\hat{a}_2)\right)$.
So, it is infeasible to construct different sub-coalitions by just reordering actions.
If someone wants to apply Shapley Value or marginal contribution to Q value, the optimal actions for sub-coalitions need to be available with a certain method.
Instead, our AMC is conducted on state value $v^H$ that is not directly related with specific actions, which is different from SQDDPG.

\subsection{The effectiveness of AMC in different goal space}
\label{app6.2}
% 2v3 -> 4v5

\begin{figure*}[h]
\subfigure[]{
\label{2v3}
\includegraphics[width=0.5\textwidth]{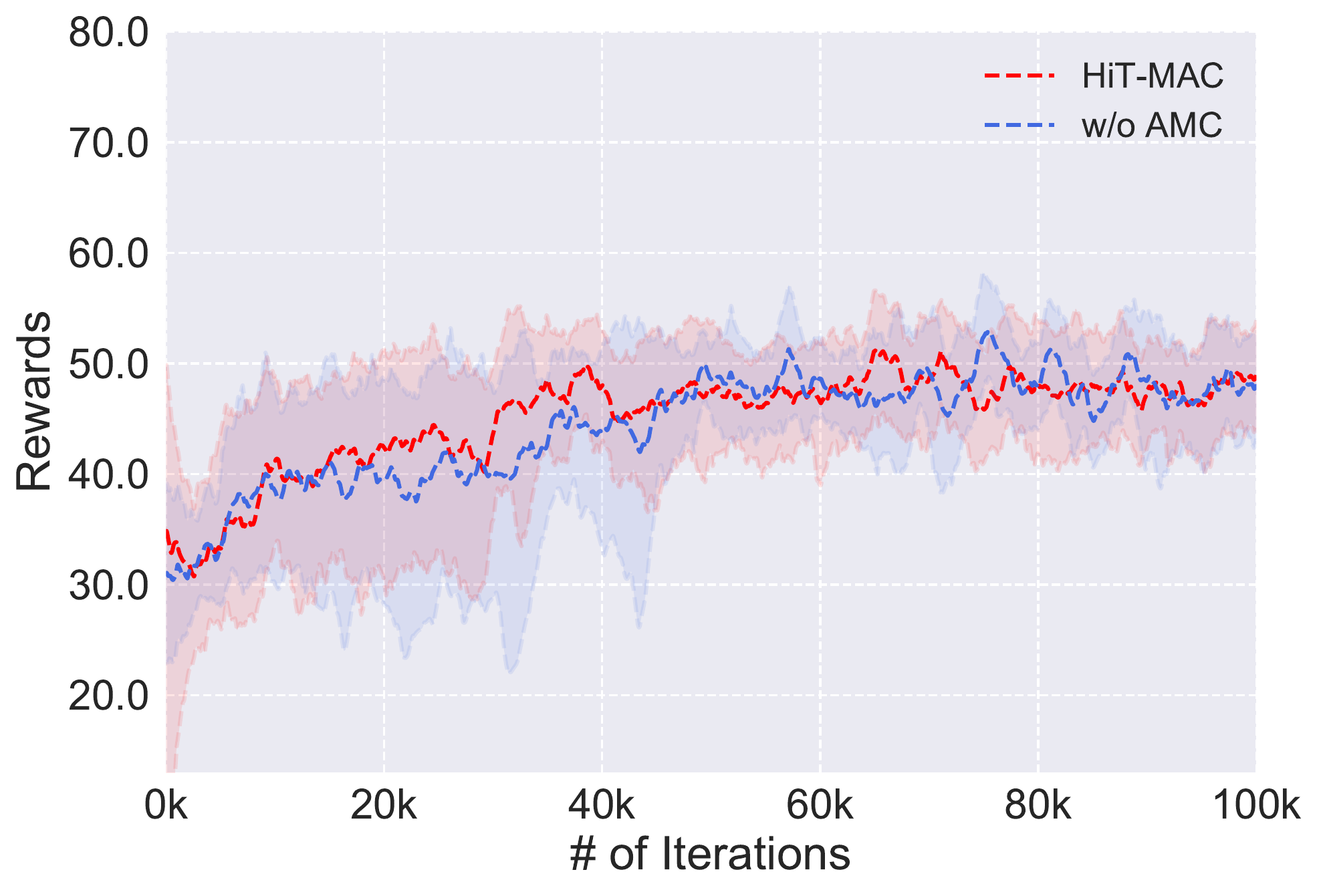}
}
\subfigure[]{
\label{4v5}
\includegraphics[width=0.5\textwidth]{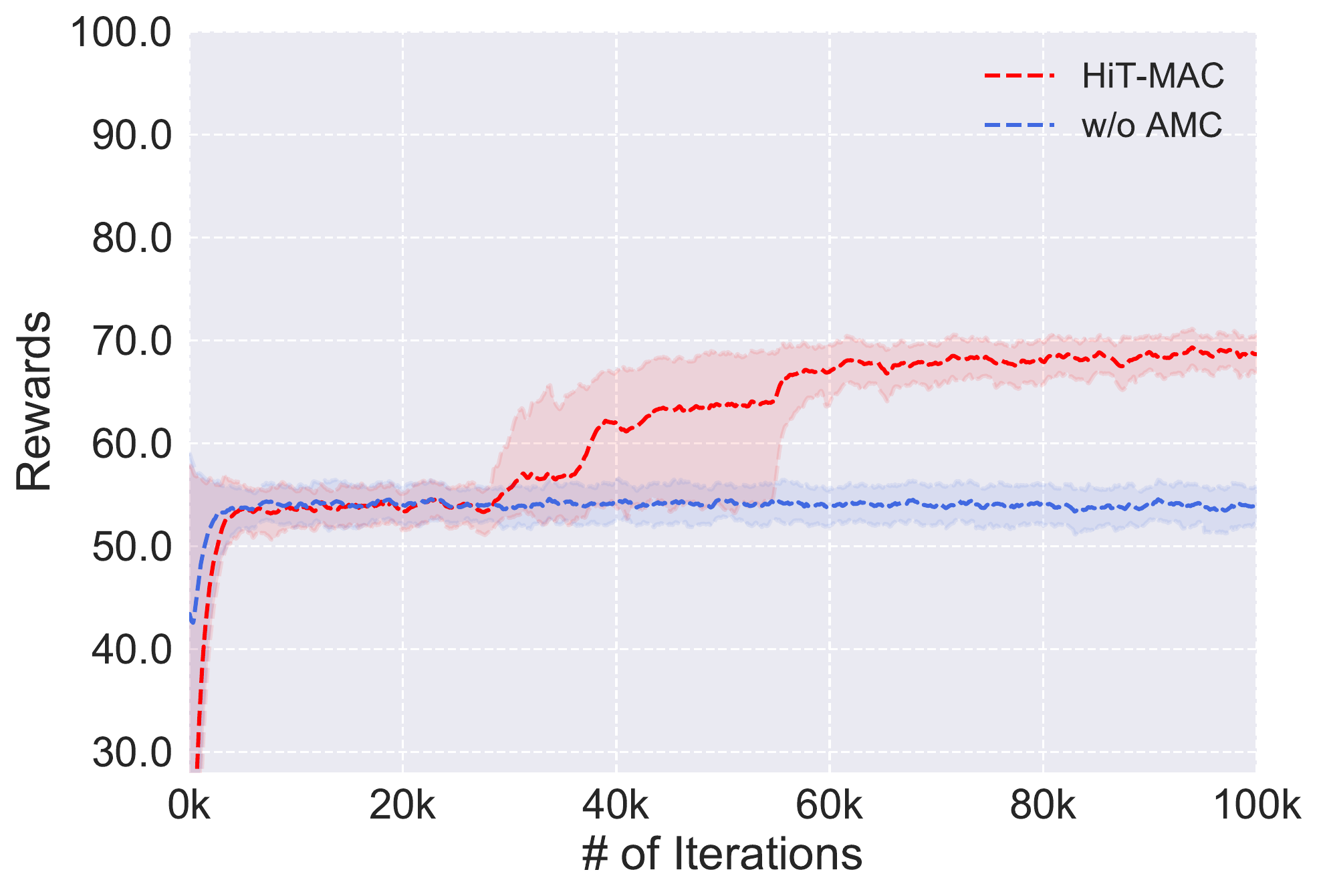}
}

\caption{
Figures prove that the effectiveness of AMC is reflected as the complexity of the environment rises, i.e. more sensors and targets. The two figures from left to right are the experiments when $n=2\&m=3$and $n=4\&m=5$, where n is the number of sensors and m is the number of the targets.}
\label{fig:learning_curve}
\end{figure*}

The results from the paper show the contribution of AMC. However, whether is AMC critical all the time? We study the situation that the contribution of Approximate Marginal Contribution(AMC) are more apparent. The studies are conducted in $n=2\&m=3$ and $n=4\&m=5$.
In the simple setting \ref{2v3}, i.e. 2 sensors cover 3 targets, we find that the one without AMC can already obtain a good performance. And the convergences are similar in this setting. 
But when the setting becomes complex, our advantages appear. When there are 4 sensors and 5 targets in the environment in Figure \ref{4v5}, the one without AMC can even not converge to a local optimal solution sometimes, since its policy entropy stay high and can not decrease. Sometimes, the one without AMC can obtain a good score, while the convergence is slower than ours. The variance between several training sessions is large.
But, our method outperforms more stably.
So, we conclude that global value estimation by AMC is effective and necessary when the cooperative setting becomes complicated.

\end{document}